Probabilistic Analogical Mapping with Semantic Relation Networks

Hongjing Lu          Nicholas Ichien          Keith J. Holyoak

University of California, Los Angeles

*Address for correspondence:*

Dr. Hongjing Lu
Department of Psychology
405 Hilgard Ave.
University of California, Los Angeles
Los Angeles, CA 90095-1563
Email: hongjing@ucla.edu

## Abstract

The human ability to flexibly reason using analogies with domain-general content depends on mechanisms for identifying relations between concepts, and for mapping concepts and their relations across analogs. Building on a recent model of how semantic relations can be learned from non-relational word embeddings, we present a new computational model of mapping between two analogs. The model adopts a Bayesian framework for probabilistic graph matching, operating on semantic relation networks constructed from distributed representations of individual concepts and of relations between concepts. Through comparisons of model predictions with human performance in a novel mapping task requiring integration of multiple relations, as well as in several classic studies, we demonstrate that the model accounts for a broad range of phenomena involving analogical mapping by both adults and children. We also show the potential for extending the model to deal with analog retrieval. Our approach demonstrates that human-like analogical mapping can emerge from comparison mechanisms applied to rich semantic representations of individual concepts and relations.

**Keywords:** analogy, relations, mapping, retrieval, distributed representations, machine learning

## Introduction

Human thinking is based not only on a vast pool of individual concepts, but also on *relations* between concepts. An explicit relation connects multiple entities, each of which fills a specific role (e.g., the relation *hit* has two roles, which might be instantiated by "hammer hits nail"). Relations greatly extend the potential to go beyond similarity of individual entities to find resemblances between situations based on *analogy*, a form of reasoning routinely used in everyday communication. Thus "hammer hits nail" is analogous to "meteor hits planet": a meteor plays the same role as a hammer, and a planet the same role as a nail, even though the corresponding entities are very dissimilar. As a more colorful example, a newspaper article describing the difficulty of using a vaccine reservation website during the COVID-19 pandemic quoted a user's complaint that, "This website is as dumb as a box of hammers, and as useful as a paper teapot" (Lopez, 2021). Such analogical metaphors call attention to important connections between dissimilar concepts so as to highlight core similarities in a creative way. Analogy plays an important role in many creative human activities, including scientific discovery (Dunbar & Klahr, 2012), engineering design (Chan & Schunn, 2015), mathematics education (Richland, Zur, & Holyoak, 2007), and metaphor comprehension (Holyoak, 2019; Holyoak & Stamenković, 2018). For reviews of relational processing in humans, see Gentner (2010), Halford, Wilson, and Phillips (2010), and Holyoak (2012); and for a review of its neural substrate see Holyoak and Monti (2021).

In general, analogical reasoning serves to transfer knowledge from a familiar and better-understood *source* analog to a more novel *target* analog. Analogical reasoning can be decomposed into multiple subprocesses (Holyoak, Novick, & Melz, 1994): *retrieval* of one or more relevant source analogs given a target, *mapping* to identify systematic correspondences between elements of a source and target, *inference* to generate new conjectures about the target based on its mapping with the source, and *schema induction* to form a more abstract representation capturing commonalities shared by the source and target. These subprocesses are interrelated, with mapping considered to be the pivotal process (Gentner, 1983). Mapping may play a role in retrieval, as mapping a target analog to multiple potential source analogs stored in memory can help identify one or more that seems promising; and the correspondences computed by mapping support subsequent inference and schema induction. Thus, because of its centrality to analogical reasoning, the present paper focuses on the process of mapping between two analogs. We also consider the possible role that mapping may play in analog retrieval.

*Computational Approaches to Analogy*

Computational models of analogy have been developed in both artificial intelligence (AI) and cognitive science over more than half a century (for a recent review and critical analysis, see Mitchell, 2021). These models differ in many ways, both in terms of basic assumptions about the constraints that define a "good" analogy for humans, and in the detailed algorithms that accomplish analogical reasoning. For our present purposes, two broad approaches can be distinguished. The first approach, which can be termed *representation matching*, combines mental representations of structured knowledge about each analog with a matching process that computes some form of relational similarity, yielding a set of correspondences between the elements of the two analogs. The structured knowledge about an analog is typically assumed to approximate the content of propositions expressed in predicate calculus; e.g., the instantiated relation "hammer hits nail" might be coded as *hit (hammer, nail)*. This type of representation requires a symbol to specify the relation *R*, a separate representation of its arguments (roles) $a_1, a_2... a_n$, and a set of bindings between relation and arguments (thereby distinguishing "hammer hits nail" from "nail hits hammer"; Halford, Wilson, & Phillips, 1998). Multiple propositions can be linked by shared arguments (Kintsch, 1988) (e.g., *hit (hammer, nail)*, *enter (nail, wall)*), or by higher-order relations (Gentner, 1983) that take one or more propositions as arguments (e.g., *cause (hit (hammer, nail), enter (nail, wall))*). Many analogy models have represented analogs using classical symbolic representations that directly correspond to predicate-calculus notation (Falkenhainer, Forbus, & Gentner, 1989; Forbus, Gentner, & Law, 1995; Forbus, Ferguson, Lovett, & Gentner, 2017; Holyoak & Thagard, 1989; Keane & Brayshaw, 1988; Thagard, Holyoak, Nelson, & Gochfeld, 1990; Winston, 1980); others have adopted specialized neural codes (e.g., tensor products or neural synchrony) that can capture both the structure and information content of propositions (Doumas, Hummel, & Sandhofer, 2008; Halford et al., 1998; Hummel & Holyoak, 1997, 2003).

Taking structured representations of individual analogs as inputs, representation-matching models use some form of similarity-based algorithm to identify correspondences. For models based on explicit symbolic representations (e.g., Falkenheiner et al., 1989; Holyoak & Thagard, 1989; Winston, 1980), analogical mapping can be viewed as a form of graph matching, where each individual analog is encoded as a structured graph with labeled nodes and edges. Mapping involves the creation of matches between elements of source and target propositions at multiple hierarchical levels (e.g., matches between objects, between relations, and between propositions). Classical symbolic representations code relations as atomic elements, which do not capture degrees of similarity (e.g., the symbol for the relation *harm* is no more similar to that for *injure* than that for *heal*). To avoid combinatorial explosion (which would arise if any element could match any other), matching of relations is typically restricted to those that are identical or closely connected in a predefined taxonomy via common superordinates (e.g., Forbus et al., 1995; Thagard et al., 1990; Winston, 1980). These restrictions limit the flexibility of classical symbolic models (Hofstadter & Mitchell, 1994). Models that express propositional content in distributed neural codes allow greater flexibility in matching relations that are similar but not identical, and can also find plausible matches between predicates with different numbers of arguments (e.g., matching a *large* animal to the *larger* of two animals; Hummel & Holyoak, 1997).

Analogy models in the broad tradition of representation matching, such as the *Structure Mapping Engine* (SME; Falkenhainer et al., 1989), which is based on classical symbolic representations; *Structured Tensor Analogical Reasoning* (STAR; Halford et al., 1998), which is based on tensor products; and *Learning and Inference with Schemas and Analogies* (Hummel & Holyoak, 1997, 2003), and the closely-related *Discovery Of Relations by Analogy* (DORA; Doumas et al., 2008), which are based on neural synchrony, capture many important aspects of human analogical reasoning. Despite their important differences, all of these models conceive of analogical reasoning in terms of a comparison process applied to complex knowledge representations designed to capture the structure of predicates and their associated bindings of entities into roles. These models, like humans, operate in a domain-general manner. They are able to solve analogies taken from stories and problems with open-ended semantic content, such as the Rutherford-Bohr analogy between planetary motion and atomic structure, or a story about military tactics that suggests an analogous solution to a medical problem (Gick & Holyoak, 1980). Also like humans, these models do not require extensive direct training on analogy problems in the target domain, and can yield what is termed "zero-shot learning": generalization to a new type of problem without prior examples of that type.

But despite their notable achievements, models based on representation matching have been handicapped by the lack of a domain-general, automated process for generating the symbolic representations required as their inputs. In principle, these representations are viewed as the products of perception (for visual analogies) or of language comprehension (for analogies between texts). But in the absence of full computational models of how either perception or comprehension might yield structured knowledge representations, the inputs to analogy models have typically been hand-coded. At a theoretical level, this limitation leads to the danger of excessive "tailorability" (Forbus et al., 2017): modelers may assume the existence of input representations that dovetail with their favored matching algorithm (e.g., positing "helpful" invariant features, identical relations, or higher-order propositions). Within circumscribed domains, significant progress has been made in automating the formation of representations suitable as inputs to SME (e.g., Lovett & Forbus, 2017; Forbus et al., 2017); and the DORA model is able to learn predicate-argument structures that provide inputs to LISA (Doumas et al., 2008; Doumas, Puebla, Martin, & Hummel, 2020). Nonetheless, representation-matching models have yet to demonstrate the ability to form structured inputs that enable analogical reasoning for open-ended domains based on perceptual or linguistic inputs. At a practical level, without automated procedures for forming the requisite representations, it is prohibitively labor-intensive to hand-code large data sets so as to enable analogical reasoning by machines.

The second major approach to computational modeling of analogy, which can be termed *end-to-end learning*, is a direct application of the type of deep learning that is at the current forefront of AI. This approach, which avoids hand-coding altogether, builds on deep neural networks that support training from raw input stimuli (e.g., image pixels, or words in a text) to a final task in an end-to-end manner. Learning in these networks is typically guided by minimizing errors in performing a particular task. This approach has moved beyond tasks involving pattern recognition (such as object classification), for which deep learning has achieved great success, to reasoning tasks. From this perspective, analogy is viewed as a task for which a deep neural network can be trained end-to-end by providing massive data consisting of analogy problems.

This approach has been applied with some success to solving visual analogies, notably problems inspired by Raven's Progressive Matrices (RPM; Raven, 1938), a variant of formal analogy problems based on matrices formed from geometric patterns. After extensive training with RPM-like problems, deep neural networks have achieved human-level performance on test problems with similar basic structure (Santoro et al., 2017; Zhang et al., 2019; Hill et al., 2019). Rather than aiming to create explicit relational representations that approximate predicate calculus, end-to-end learning forms representations consisting of complex conjunctions of features distributed across a multilayer network. There is no separable process of assessing the similarity of the two analogs. Rather, deep learning creates representational layers culminating in a final decision layer that selects or generates the best analogical completion. That is, learned representations of analogs are directly linked to the task structure in which they are used.

End-to-end learning models represent the current highwater mark in automated analogical inference, as hand-coding of inputs is entirely avoided. However, these AI systems appear quite implausible if interpreted as psychological models. First, their success depends on datasets of massive numbers of RPM-like problems (e.g., 1.42 million problems in the PGM dataset, Barrett et al., 2018; and 70,000 problems in the RAVEN dataset, Zhang et al., 2019). For example, Zhang et al. (2019) used 21,000 training problems from the RAVEN dataset, and 300,000 from the PGM dataset. This dependency on direct training in a reasoning task using big data makes the end-to-end learning approach fundamentally different from human analogical reasoning. When the RPM task is administered to a person, "training" is limited to general task instructions. Because the task is intended to provide a measure of fluid intelligence—the ability to manipulate *novel* information in working memory (Snow, Kyllonen, & Marshalek, 1984)—extensive pretraining on RPM problems defeats the entire purpose of the test.

Second, the generalization ability of current end-to-end learning models is limited to test problems that are very similar in content and structure to the training problems. If the content of analogy problems deviates even modestly from that used in the training examples, generalization falls well short of human performance (e.g., Ichien et al., 2021). The end-to-end approach thus fails to account for the human ability to achieve zero-shot learning by analogical transfer. Arguably, this shortcoming is directly related to the fact that end-to-end deep learning does not create explicit relational representations (Doumas et al., 2020).

*Eduction of Relations*

In the present paper we describe a novel computational model of analogical mapping that addresses the basic question of how inputs to the reasoning process can be generated. Our model is in the tradition of representation matching (making central use of graph matching), but differs from previous proposals in its approach to relation representation. Our approach is not rooted in the logic of predicate calculus, but rather in the seminal theories of human intelligence formulated a century ago. Consider a simple verbal analogy in the proportional format (*A:B* :: *C:D*) often used on intelligence tests, e.g., *hot* : *cold* :: *love* : *hate.* We can think of the *A:B* pair as the source and *C:D* as the target. The first thing to note is that the problem statement does not specify any relations. Rather, as Charles Spearman observed, the initial step in solving the analogy is to perform what he termed the *eduction of relations*: in his own (rather awkward) words, "*The mentally presenting of any two or more characters* (simple or complex) *tends to evoke the knowing of relation between them*" (Spearman, 1923, p. 63; italics in original). That is, the reasoner must first mentally "fill in the blanks" in the problem as posed, by retrieving or computing the relation between *A* and *B*, and that between *C* and *D*. Once these relations have been educed, the reasoner can perform a second basic step, the *eduction of correlates*: assessing the similarity of the *A:B* and *C:D* relations to determine whether they are analogous.

For verbal problems, Spearman's concept of "relation" refers to the semantic relation between concepts denoted by words. Semantic relations are more than mere associations; e.g., *hot* : *cold* :: *love* : *adore* consists of two word pairs that are each strongly associated via a salient relation, but the problem does not form a valid analogy because the *A:B* and *C:D* relations mismatch. At the same time, semantic relations do not necessarily correspond in a direct way to "predicates" as typically incorporated into analogy models. The canonical examples of relations as predicates center on verbs (and other linking words), as in *hit (hammer, nail).* But at the level of semantic relations, one can represent "*hammer hits nail*" by identifying relations for the three pairs of content words: *hammer* : *hit*, *hit* : *nail*, *hammer* : *nail*. Verbs, like nouns, denote concepts that enter into pairwise semantic relations—they are not the semantic relations themselves. In the present paper we will refer to verbs and similar linking words as "predicates" when we wish to distinguish such words from semantic relations.

Proportional analogies were once a focus of psychological work on analogy (e.g., Sternberg, 1977), but fell into disfavor. More recent models in the tradition of representation matching have bypassed proportional analogies on the grounds they are too simplistic—they apparently require just the matching of single relations, rather than finding a rich mapping between *systems* of relations (Gentner, 1983). Indeed, proportional analogies do not require the reasoner to perform a mapping process at all: the format directly specifies the correspondences ($A \rightarrow C$, $B \rightarrow D$), and validity depends solely on the similarity of the educed *A:B* and *C:D* relations. Yet paradoxically, these "simplistic" analogies pose a basic problem for the sophisticated computational models that can deal with analogies between stories and word problems. The computational models of recent decades require relation-centered propositions as inputs—which is exactly what proportional analogies do not provide. Models that have addressed proportional analogies have simply assumed that relations between word pairs are prestored in long-term memory, ready to be retrieved (e.g., Morrison et al., 2004). For a simple case such as *hot* : *cold* :: *love* : *hate*, it is indeed plausible that people have prestored the relevant relation, *opposite-of*. But people can also solve analogies based on less familiar relations, as in *mask* : *face* :: *alias* : *name*. In such cases reasoners may not have considered the relations between the word pairs prior to receiving the analogy problem. Rather, relations must be educed from representations of the concepts being related. A model that accomplishes the eduction of relations between paired concepts would at least partially address the problem of how relational representations of analogs can be formed by an autonomous reasoner, reducing the need for hand-coding by the modeler. However, modeling the eduction of relations presupposes finding an answer to a yet more basic question: how are semantic relations acquired in the first place?

*Plan of the Paper*

In the remainder of this paper we present and test a new model of analogical mapping over a range of verbal reasoning tasks varying in complexity. The model operates on *semantic relation networks*—graphs in which feature vectors capture the rich semantics both of individual concepts (nodes in graphs) and of pairwise relations between concepts (edges). The proposed mapping model serves as a module in a broader system, making use of the outputs of additional modules that address the acquisition and eduction of relations, as well as text processing. In the spirit of other recent computational models of human cognition (e.g., Battleday, Peterson, & Griffiths, 2020), we build on work integrating developments in deep learning with theoretical ideas from cognitive science.

The inputs to the mapping model are two sets of concepts, respectively selected from the source and target analogs. For analogies based on texts, we explore the potential for using AI algorithms for natural language processing (NLP) to aid in selection of key concepts. The model adopts rich semantic representations (embeddings) for individual concepts, partially automates the creation of skeletal relational structure for analogs, and then applies Bayesian probabilistic inference to find correspondences between key concepts in each analog by maximizing the similarity between two analogs under the constraint of favoring isomorphic mappings. Mapping in the proposed model depends on semantic relations of the sort considered by Spearman (1923), but does *not* require, nor directly operate on, complex hierarchies of propositions. The model is domain-general, and does not require explicit training in solving analogy problems within any particular domain. The aim is to capture the power of representation matching to produce zero-shot learning by analogy, while at the same time pursuing the theoretical goal of end-to-end learning: to automate the creation of representations that provide the proximal inputs to analogical mapping.

Because the overall model is modular in nature, some components could readily be altered or replaced. In dealing with issues related to text processing, we make use of NLP algorithms that have proven helpful in work to date, but these are clearly imperfect and not intended to be definitive. The module that creates vector representations of word meaning might be replaced by some other machine-learning algorithm. The module we use to create vector representations of relation meanings is also subject to revision. Here we compare one model (an extension of our own earlier work) with an alternative baseline model, as well as with additional variants created by systematic ablations (see Supplemental Information). The central contribution of the present paper is the proposed mapping module, which takes vector representations of concepts and relations as inputs and yields analogical correspondences as outputs. This mapping model would operate in essentially the same manner if the modules that create its inputs were varied.

The scope of the model as presented here is limited to verbal analogies (although a similar approach may be applicable to visual analogies; Ichien et al., 2021). We believe the mapping model could in principle be instantiated in a neural network; however, to maximize its generality we provide a Bayesian formulation based on probabilistic inference. We will first describe the creation of inputs to the mapping model: vector representations of individual concepts, and of semantic relations between concepts. We then focus on the mapping model itself. We apply the model to a novel analogy task that requires the eduction and integration of multiple semantic relations, as well as to a series of classic experiments drawn from the analogy literature. In addition to analogical mapping, we consider how the model could be applied to the problem of analog retrieval. Our treatment of psychological phenomena is selective, emphasizing basic findings regarding human judgments of preferred mappings between analogs, and propensities to retrieve different types of source analogs in response to a given target analog.

### Forming Representations of Word Meanings and Semantic Relations

A general system for analogical mapping that takes verbal inputs must accomplish four component tasks: (1) creating representations of word meanings; (2) learning and then recognizing semantic relations between words; (3) integrating representations of word meanings and semantic relations to code complete analogs; and (4) comparing analogs to generate a set of correspondences between them. Figure 1 schematizes a set of individual models that operate as modules to support and perform analogical mapping, applied to a simple example. The first two tasks are accomplished by versions of existing models; the latter two tasks are performed by a new model introduced in the present paper.

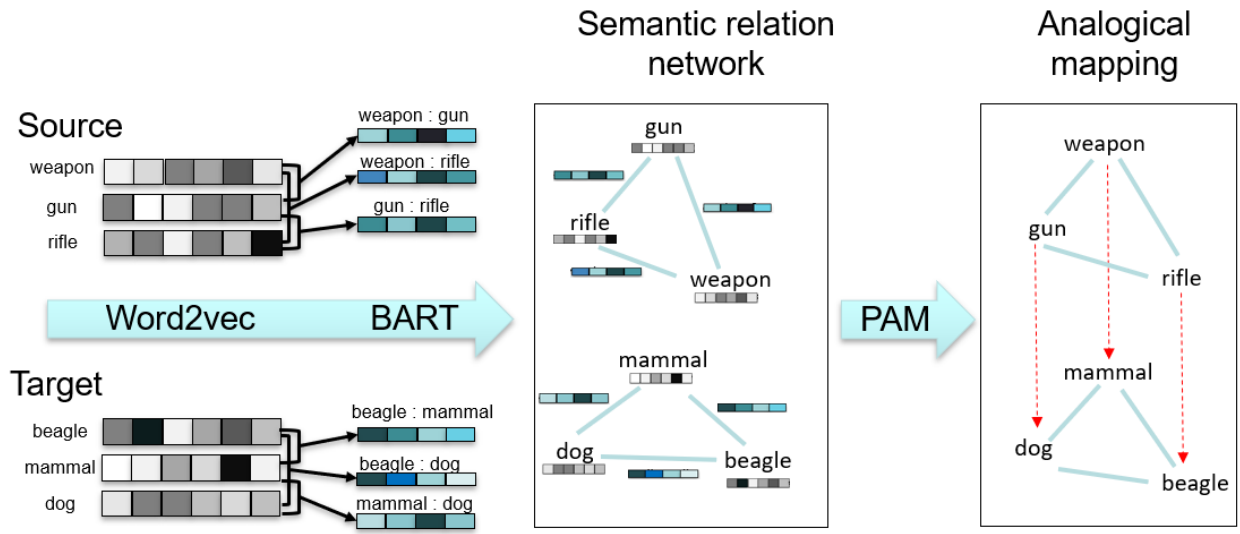

*Figure 1.* An illustration of analogical reasoning based on semantic mapping, using Category triplets as an example. From left: Word embeddings (provided by Word2vec) are obtained to represent the semantic meaning for each concept (keyword); relation vectors (from BART) are obtained to represent semantic relations instantiated for pairs of concepts. Shades in the blocks represent different values in vectors. Middle: Unaligned semantic relation networks, in which nodes are individual concepts and edges are semantic relations between concept pairs, are created for each analog. Word embeddings (illustrated as gray blocks) are assigned as node attributes and relation vectors (illustrated as blue blocks) are assigned as edge attributes. Right: Aligned semantic relation networks are generated by performing probabilistic analogical mapping (with PAM) to find mappings between the concepts in source and target that maximize the combined similarity based on keywords and relations.

### Creating Representations of Word Semantic Meanings Using Word2vec

As the first step toward automating analogical mapping, we adopt semantic representations of individual words generated by a machine-learning model, Word2vec (Mikolov et al., 2013). Word2vec and similar models based on distributional semantics, such as GloVe (Pennington, Socher, & Manning, 2014) and BERT (Devlin, Chang, Lee, & Toutanova, 2019), have proved successful in predicting behavioral judgments of lexical similarity or association (Hill, Reichart, & Korhonen, 2015; Hofmann et al., 2018; Pereira, Gershman, Ritter, & Botvinick, 2016; Richie & Bhatia, 2021), neural responses to word and relation meanings (Huth, de Heer, Griffiths,

Theunissen, & Gallant, 2016; Pereira et al., 2018; Zhang, Han, Worth, & Liu, 2020), and high-level inferences including assessments of probability (Bhatia, 2017; Bhatia, Richie, & Zou, 2019) and semantic verification (Bhatia & Richie, in press). In the simulations reported here, the semantic meanings of individual concepts are represented by 300-dimensional embeddings created by Word2vec after training on a corpus of articles drawn from Google News.

*Creating Representations of Relations between Concepts using BART*

The second major component of the overall system is a model that acquires representations of the semantic relations between concept words. In keeping with the use of embeddings to represent individual word meanings, we represent relations as vectors. Once representations of semantic relations have been created, then in principle it becomes possible to solve proportional verbal analogies by computing the similarity of the *A:B* relation to the *C:D* relation by some generic measure, such as cosine similarity. Word2vec itself has been applied to four-term verbal analogies by computing the cosine distance between difference vectors for *A:B* and *C:D* pairs, a measure we refer to as *Word2vec-diff* (Zhila, Yih, Meek, Zweig, & Mikolov, 2013; for a different distributional approach based specifically on relation terms, see Turney, 2008, 2013). Although direct application of Word2vec achieved some success for analogies based on semantically-close concepts, it fails to reliably solve problems based on more dissimilar concepts (Linzen, 2016; Peterson, Chen, & Griffiths, 2020). In the present paper we use Word2vec-diff as a baseline model, in which relations are coded in a generic fashion simply as difference vectors.

To move beyond generic difference vectors, Word2vec vectors for pairs of individual words can be used as inputs to learn representations of relations in a transformed semantic relation space. *Bayesian Analogy with Relational Transformations* (BART) (Lu, Chen, & Holyoak, 2012; Chen, Lu, & Holyoak, 2017; Lu, Wu, & Holyoak, 2019), using supervised training with concatenated word pairs coded by Word2vec embeddings, can learn to estimate the probability that any pair of words instantiates any abstract semantic relation drawn from a pool of such relations. This pool includes 135 abstract relation categories, such as *category* (*fruit* : *apple*), *similar* (*house* : *home*), *contrast* (*hot* : *cold*), *part-whole* (*finger* : *hand*), and *case relation* (*read* : *book*). Semantic relations between words are then coded by BART as distributed representations over its set of learned abstract relations. After learning, BART calculates a relation vector consisting of the posterior probability that a word pair instantiates each of the learned relations.

 *Learning in BART*. The basic operation of the BART model (described in detail by Lu et al., 2019) is illustrated in Figure 2. BART uses a three-stage process to learn semantic relations from nonrelational inputs consisting of positive and negative examples of each target relation (typically about 20 positive and 70 negative examples). The initial input (bottom layer of the network sketched in Figure 2, left) consists of a concatenated vector of length 600 representing a pair of words (where each word in a pair is coded by a 300-dimension Word2vec embedding). In its first stage, the model augments this raw feature vector by (a) computing the difference in the value of each feature between the two words in a pair, (b) ordering these differences by magnitude, and (c) creating 600 additional features consisting of the raw features reordered according to difference magnitudes. These ranked features will differ for each word pair used in training. Augmenting the raw semantic features with ranked features partially mitigates the problem that across instances, different semantic features may be relevant to a relation (e.g., *love* : *hate* involves features related to emotion, whereas *rich* : *poor* involves features related to wealth); and worse, the features of word embeddings are typically "entangled" (i.e., individual features are not readily interpretable). Difference ranking places features that generate differences of similar magnitude (and hence are relatively likely to serve similar semantic functions) into correspondence, without assuming any prior knowledge about which individual features are relevant to any relation for any particular word pair. This first stage culminates in the generation of a 1,200-dimension augmented feature vector for each word pair, consisting of the concatenation of raw and ranked feature vectors for each word in the pair (second layer from bottom in Figure 2, left).

In its second stage, BART applies logistic regression with elastic net regularization to difference vectors for all features, selecting a subset of features in the second layer that are most statistically important in predicting the relation being trained (yielding the third layer from bottom in Figure 2, left), and estimating the associated coefficients. In its third stage, BART uses Bayesian logistic regression to estimate the weight distribution representing the target relation $R$ based on the selected features of word pairs for all training examples. This regression includes a contrast prior derived from the second stage (i.e., for each feature included in stage 3, the initial coefficient for the first word in a pair is set equal to that estimated in stage 2, and the initial coefficient for the second word is set to the negative of that value).

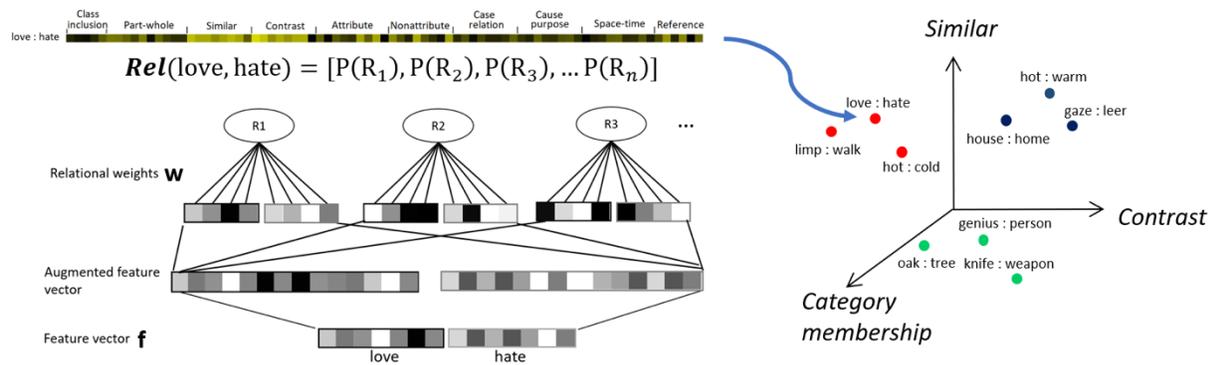

*Figure 2*. Left: Schematic illustration of BART model architecture for relation representation. The bottom layer of the BART model is a concatenated input vector based on the two words in a pair; the top layer indicates the set of learned relations (ellipses indicate additional relations beyond the three illustrated here). After learning, the semantic relation between any two words is represented as a vector of the posterior probabilities of each learned relation; the relation vector (Rel) linking *love* and *hate* is shown on the top as an illustration. Right: Semantic relations formed by BART generate a transformed (and disentangled) space in which pairs instantiating similar sets of relations tend to show similar patterns in relation vectors, and hence are located close to one another in the relation space.

*Relation vectors in BART*. As illustrated in Figure 2 (right), BART effectively re-represents the relation between two specific concepts as a vector in a new semantic space (for a related approach see Roads & Love, 2021), thereby educing the relation between any pair of words. The specific relation between any two words is thus coded as a distributed representation. This representation is disentangled in that each element in the relation vector corresponds to the posterior probability that a particular meaningful relation holds between the concepts. BART's distributed representations enable the model to generalize to new word pairs that may be linked by relations on which the model had not been specifically trained. By comparing the similarity between relation vectors (assessed by cosine distance), semantic relation representations derived by BART have been used to solve verbal analogies in *A:B :: C:D* format (Lu et al., 2019), to predict human judgments of relation typicality and similarity (Ichien, Lu, & Holyoak, 2021), and to predict patterns of similarity in neural responses to relations during analogical reasoning (Chiang, Peng, Lu, Holyoak, & Monti, 2021).

For the present project, we trained BART by combining two datasets of semantic relations. The first dataset included 79 specific relations from a taxonomy of ten abstract semantic relations (Bejar, Chaffin, & Embretson, 1991; Jurgens, Turney, Mohammad, & Holyoak, 2012), each with at least 20 word pairs instantiating the same relation. The second dataset (Popov, Hristova, & Anders, 2017) provided 56 additional specific relations, each with 12-25 word pairs instantiating the same relation. For each of 135 relations, BART learns to select semantic features $\mathbf{f'}_L$ and infer the weight distributions $\mathbf{w}$ associated with semantic features for each relation from training data $(\mathbf{f'}_L, R_L)$ using variational Bayesian methods, as summarized in the previous section. After learning, BART can estimate how likely a word pair $\langle \boldsymbol{f_1}, \boldsymbol{f_2} \rangle$ instantiates a particular relation $R_i$ using the computation,

$$P(R_i = 1|\boldsymbol{f_1}, \boldsymbol{f_2}) = \int P(R_i = 1|\boldsymbol{f_1}, \boldsymbol{f_2}, \mathbf{w})P(\mathbf{w}|\mathbf{f'}_L, R_L)d\mathbf{w}. \qquad (1)$$

BART then calculates the posterior probability that the word pair instantiates each of the relations in its pool of 270 learned relations, resulting in a distributed representation of relation vector between two words, $\boldsymbol{Rel}(\boldsymbol{f_1}, \boldsymbol{f_2}) = \langle P(R_1 = 1|\boldsymbol{f_1}, \boldsymbol{f_2}), \ P(R_2 = 1|\boldsymbol{f_1}, \boldsymbol{f_2}), \dots, P(R_k = 1|\boldsymbol{f_1}, \boldsymbol{f_2}) \rangle$. In addition, BART automatically forms representations for the converse of each trained relation. For example, after learning relation representations for the category-instance relation, BART can generate representations for instance-category relation by swapping the weights attached to features of the two words in a pair. Hence, BART relation vectors include 270 dimensions, each encoding posterior probability of a word pair instantiating a relation in a repertoire of 270 relations.

Ichien et al. (2021) found that in modeling human judgments of relational similarity, BART's predictions are improved by applying a nonlinear power transformation (with the power parameter of 5) to the relation vector. This transformation emphasizes the contributions of those relations with higher posterior probabilities in the similarity calculation ("winners take most"). In modeling mapping of Category triplets (Simulation 1), a parameter search confirmed that a power of 5 yielded the best fit. According, all BART vectors used in simulations reported here include this power transformation.

*Role Vectors in BART.* Analogical mapping goes beyond judgments of relation similarity in that a coherent mapping requires not only that individual matched relations be similar to one another, but also that elements play corresponding roles across multiple relations (i.e., mapping is sensitive to the bindings of concepts into relational roles). It is clear that humans are sensitive to semantic roles in relation processing. For example, people not only recognize that *mammal* : *dog* instantiates the relation *category-instance*, but also that *mammal* plays the first role (*category*) rather than the second role (*instance*). Hence, role information is linked to relation representations, as assumed by some previous computational models of analogy (e.g., Hummel & Holyoak, 1997). In addition to evaluating an overall relation, humans are able to evaluate how well entities fill specific roles in that relation (Popov, Pavlova, & Hristova, 2020). Markman and Stilwell (2001) provided a taxonomy of categories that explicitly distinguishes between relational and role-governed categories. A study by Goldwater, Markman, and Stilwell (2011, Experiment 3) demonstrated that learning a novel relational structure (instantiated as a novel verb; e.g., learning that *to cake* means "to make a cake") licenses learning of novel roles (e.g., a *caker* is "someone who cakes"). This path of acquisition is consistent with how BART acquires roles: relations are learned first, then roles are extracted from the learned relations. (In contrast, the DORA model of Doumas et al., 2008, makes the opposite assumption, that individual roles are acquired first and then combined to form multi-place predicates.) Although people sometimes detect role-based categories without explicit instruction (Goldwater, Bainbridge, & Murphy, 2016), category labels and analogical comparisons increase general sensitivity to role-based categories (Goldwater & Markman, 2011). Moreover, objects occupying the same role in a relation (e.g., *predator*) come to be viewed as more similar to each other overall (Jones & Love, 2008).

In order to represent bindings of concepts to roles in semantic relations, which is required to compute systematic mappings, the version of BART used in the present paper introduces a new extension: the model includes learned representations of the relational roles played by individual concepts. BART relation vectors were augmented with role vectors indicating the probability that the first word in a pair fills the first role of the relation. (Because the relevant probabilities must sum to 1 across the two words in a pair, it is sufficient to explicitly represent the probability for only the first word.) These role vectors were created using the same training data that was used to train relations in BART. The first word in each word pair instantiating a relation was treated as a positive example, and the second word was treated as a negative example.

Role learning operates on top of BART's relation learning. Weighted feature inputs are generated by elementwise product , $\mathbf{w} \circ \boldsymbol{f}$ , of the semantic features $\boldsymbol{f}$ selected by BART and its learned relation weights $\mathbf{w}$ in its third stage connecting the top two layers in Figure 2. Taking weighted feature vectors derived from BART as the inputs, the model's Bayesian logistic regression algorithm is reapplied to learn weight distributions $\boldsymbol{\omega}$ for role representations from training data $< \mathbf{f}'_L, R_L >$. After learning, the role-based weight distributions of $\boldsymbol{\omega}$ are used to estimate the posterior probability that the first word $\boldsymbol{f_1}$ in a pair plays the first role in the $i$th relation:

$$P(r_i = 1|\mathbf{w} \circ \boldsymbol{f_1}) = \int P(r_i = 1|\mathbf{w} \circ \boldsymbol{f_1}, \boldsymbol{\omega}) P(\boldsymbol{\omega}|\mathbf{w} \circ \mathbf{f}'_L, R_L) d\boldsymbol{\omega}. \qquad (2)$$

In the remainder of the paper we use the term "relation vector" as shorthand for BART's concatenation of relational and role vectors. The final relation vectors used in the work reported here consisted of 540 dimensions: 270 posterior probabilities of a word pair instantiating each of the 270 relations BART has acquired, concatenated with 270 posterior probabilities that the first word in the pair plays the first role for each of the 270 relations. (Performance comparisons with models based on reduced vectors are provided in Supplemental Information.)

## Probabilistic Analogical Mapping (PAM)

Using representations of individual concepts generated by Word2vec and semantic relations between pairs of concepts generated by BART, we have developed a model to accomplish the third and fourth tasks noted above: forming representations of entire analogs and then computing a mapping between them. A general model of analogical mapping that takes two complex analogs as inputs must capture the human ability to integrate multiple relations (Gentner, 1983; Halford et al., 1998). In the example shown in Figure 1, an ordered sequence of three category concepts (*weapon* : *gun* : *rifle*) is mapped to a set of scrambled concepts from a different domain (*dog*, *beagle*, and *mammal*). When the source and target analogs involve multiple pairwise relations, as in this example, inherent mapping ambiguities may arise. For example, *weapon* : *gun* considered alone could map to either *mammal* : *dog* or *dog* : *beagle*, because all of these pairs instantiate the *superordinate-of* relation. As we will show in an experiment reported below, humans can reliably solve such analogy problems; a comparable requirement to integrate multiple relations arises in many other relational reasoning paradigms, such as transitive inference (Halford et al., 1998; Waltz et al., 1999). To resolve ambiguity in local mappings, a reliable analogy model must assess relation similarities and integrate across relations based on mapping constraints.

To compute mappings between concepts in analogies involving multiple relations, we have developed a domain-general model of *Probabilistic Analogical Mapping* (PAM). The model combines graph matching based on an algorithm that performs constraint satisfaction (similar in spirit to comparison models such as Holyoak & Thagard's ACME, 1989, Goldstone's SIAM, 1994, and Goldstone & Rogosky's ABSURDIST, 2002) with vector representations of both concepts and relations. PAM operates on *semantic relation networks*, a type of graph structure in which nodes represent individual concepts and edges represent semantic relations between concepts. (Note that "concepts" may include nouns, verbs, adjectives and other predicates.) The semantic relation networks created for each analog (Figure 1, middle) have the form of *attributed graphs* (in the terminology of graph matching; Gold & Rangarajan, 1996), because nodes and edges are assigned numerical *attributes,* capturing the semantic meanings of individual concepts and their pairwise relations. The attribute for each node is the Word2vec embedding of a key concept word, and the attribute for each edge is the corresponding relation/role vector generated by BART (i.e., concepts and relations are represented in separate feature spaces). Once key concepts have been specified, semantic relation networks representing individual analogs are created in an automated fashion, without hand-coding of either concept meanings or semantic relations.

*Mapping Based on Semantic Relation Networks*

Using the semantic relation networks created for the source and target analog, PAM performs analogical mapping using a probabilistic approach (Gold & Rangarajan, 1996). Source and target analogs can be represented as two graphs of semantic relation networks, $g$ and $g'$, respectively. A semantic relation network for the source analog is defined as an attributed graph $<N, E, A>$ where each node $N$ and each edge $E$ is assigned an attribute $A$. As applied to verbal analogies, nodes are words for individual concepts and edges are semantic relations between words. Let $i$ and $j$ be indices of nodes in the graph. $A_{ii}$ indicates the semantic attribute of the $i$th concept, and $A_{ij}$ indicates the relation attribute of the edge between the $i$th concept and the $j$th concept. The target analog can be represented as graph $g'$ with $i'$ and $j'$ as indices of nodes in the graph. $M_{ii'} = 1$ if the $i$th concept node in the source analog maps to the $i'$th node in the target analog, and $M_{ii'} = 0$ if the two concepts are not mapped. The goal of the model is to estimate the probabilistic mapping matrix $\boldsymbol{m}$, consisting of elements denoting the probability that the $i$th node in the source analog maps to the $i'$th node in the target analog, $m_{ii'} = P(M_{ii'} = 1)$. Using a Bayesian approach, given two semantic relation networks, PAM aims to infer a mapping $\boldsymbol{m}$ between concepts in the two analogs so as to maximize its posterior probability with the constraints $\forall i \sum_{i'} m_{ii'} = 1, \ \forall i' \sum_{i} m_{ii'} = 1$:

$$P(\boldsymbol{m}|g, g') \propto P(g, g'|\boldsymbol{m})P(\boldsymbol{m}). \tag{3}$$

The likelihood term $P(g, g'|\boldsymbol{m})$ is determined by the semantic similarity between concepts and relations weighted by mapping probability. We define the log-likelihood as

$$\log\big(P(g, g'|\boldsymbol{m})\big) = \sum_i \sum_{j \neq i} \sum_{i'} \sum_{j' \neq i'} m_{ii'} m_{jj'} S\big(A_{ij}, A_{i'j'}\big) + \alpha \sum_i \sum_{i'} m_{ii'} S(A_{ii}, A_{i'i'}), \quad (4)$$

where $S\big(A_{ij}, A_{i'j'}\big)$ represents the normalized relation similarity between edge attributes of the relation instantiated by the $i$th and $j$th concepts in one analog and that instantiated by the $i'$th and $j'$th concepts in the other analog, with the constraints of $\sum_{i',j'} S\big(A_{ij}, A_{i'j'}\big) = 1$ and $\sum_{i,j} S\big(A_{ij}, A_{i'j'}\big) = 1$. $S(A_{ii}, A_{i'i'})$ represents the normalized similarity between node attributes of individual concepts.

The normalization of similarities were implemented using a bistochastic normalization procedure developed by Cour et al. (2006). The goal of this normalization procedure is to selectively weight the influences of particular concepts and relations on mapping. Intuitively, this normalization operation decreases the influence of concepts (nodes) and relations (edges) that are not discriminative (e.g., those showing indistinguishable similarity to many concepts/relations in the other analog), and correspondingly increases the influences of discriminative concepts and relations (those showing high similarity scores to a small number of concepts/relations but low similarity scores to other concept/relations in the other analog). All similarity scores were calculated using cosine similarity. The parameter $\alpha$ in Equation 4 is a weighting parameter that controls the relative importance of lexical similarity (nodes) versus relation similarity (edges) on mapping, with higher values indicative of greater emphasis on entity-based lexical similarity in comparison to relation similarity. The $\alpha$ parameter allows PAM to capture psychological evidence that a variety of factors can alter human sensitivity to relation versus entity-based similarity (Goldstone, Medin, & Gentner, 1991; Markman & Gentner, 1993; Vendetti, Wu, & Holyoak, 2014). The fundamental assumption is that concepts (nodes) and relations (edges) constitute two separable pools of semantic information (entity-based and relation-based) that jointly drive judgments of similarity between analogs.

The prior term in Equation 3 captures generic constraints that higher prior probability is assigned to deterministic mappings with the constraints $\forall i \; \sum_{i'} m_{ii'} = 1, \; \forall i' \; \sum_{i} m_{ii'} = 1$, and is defined with a parameter $\beta$ to control the strength of the prior as,

$$P(\boldsymbol{m}) = e^{\frac{1}{\beta}\sum_{i}\sum_{i'} m_{ii'} \log m_{ii'}}. \tag{5}$$

To implement the inference in Equation 3, we employ a graduated assignment algorithm (Gold & Rangarajan, 1996), variants of which have been applied to matching problems in computer vision (Lu & Yuille, 2005; Menke & Yang, 2020). The algorithm incorporates soft assignments in graph matching. A deterministic one-to-one correspondence constraint requires that a node in one graph must match to one node in the other graph and vice versa, with the mapping values $\boldsymbol{m}$ either 0 or 1. The graduated assignment algorithm relaxes this constraint by allowing probabilistic mapping values that lie in the continuous range [0, 1]. The matching algorithm minimizes the energy function (equivalent to maximizing the posterior probability defined in Equation 3) with respect to the matching matrix:

$$\boldsymbol{E}[\boldsymbol{m}] = -\sum_{i,i',j,j'} m_{ii'} m_{jj'} S(A_{ij}, A_{i'j'}) - \alpha \sum_{i,i'} m_{ii'} S(A_{ii}, A_{i'i'}) - \frac{1}{\beta}\sum_{i}\sum_{i'} m_{ii'} \log m_{ii'} \,,$$

$$s.t. \; \forall i \; \sum_{i'} m_{ii'} = 1, \; \forall i' \; \sum_{i} m_{ii'} = 1. \tag{6}$$

where $\alpha$ controls the relative weights between lexical similarity of concepts (node attributes) and relational similarity (edge attributes). $\beta$ is a control parameter used to slowly push the values of mapping variables toward either 0 or 1 by applying the softmax function through iterations. An annealing operation with normalization is implemented to gradually increase $\beta$ over iterations to approximate the one-to-one constraint. In the implementation code, we used a fixed number of iterations (500) for all simulations, which was reliably sufficient to ensure stable mapping solutions. Table 1 provides pseudo-code for the probabilistic mapping algorithm.

*Table 1*. Pseudo-code for Probabilistic Analogical Mapping (PAM) algorithm.

Compute semantic similarity of nodes and relation similarity of edges
Apply bistochastic normalization to similarity matrix

$\beta \leftarrow \beta_0$
$\boldsymbol{m} \leftarrow$ equal probability to match to all concepts
**for** iteration **do**
    compute compatibility matrix based on node/edge similarities and mapping:
    $\forall i \in G_1, \forall i' \in G_2$
    $Q_{ii'} \leftarrow \sum_i \sum_{j \neq i} \sum_{i'} \sum_{j' \neq i'} m_{ii'} m_{jj'} S(A_{ij}, A_{i'j'}) + \alpha \sum_i \sum_{i'} m_{ii'} S(A_{ii}, A_{i'i'})$

    update soft assignments:
    $\forall i \in G_1, \forall i' \in G_2$
    $m_{ii'} \leftarrow e^{\beta Q_{ii'}}$

    update mapping matrix *m* by normalizing across all rows:
    $\forall i \in G_1, \forall i' \in G_2$
    $m_{ii'} \leftarrow \frac{m_{ii'}}{\sum_j m_{ji'}}$

    update mapping matrix *m* by normalizing across all columns:
    $\forall i \in G_1, \forall i' \in G_2$
    $m_{ii'} \leftarrow \frac{m_{ii'}}{\sum_{j'} m_{ij'}}$

    $\beta \leftarrow \beta + \frac{\beta_0}{10}$
**end**

The soft constraints incorporated in PAM are closely related to those specified in the multiconstraint theory of analogical mapping, which was first instantiated in the *Analogical Constraint Mapping Engine* (ACME; Holyoak & Thagard, 1989, 1995). Mappings between similar concepts are favored based on greater semantic similarity of node attributes, and mappings of similar relations are favored based on greater relational similarity of edge attributes. Nodes and edges can have varying importance weights, reflecting greater attention to elements important for the analogist's goals. The preference to move mapping variables toward 0 or 1 with normalization implements a soft assignment of one-to-one mappings between concepts across analogs (favoring isomorphic mappings).

*Incorporating Relation Constraints Provided by Texts*

A longstanding (though unrealized) goal for models of analogy has been to enable reasoning based on text inputs (e.g., Winston, 1980). When given text inputs (rather than analogies based simply on sets of words), PAM makes use of constraints provided by sentence structure. In forming semantic network graphs, PAM can naturally accommodate relation constraints provided by textual descriptions of analogs, by controlling the presence and absence of relation links and their directionality. Because semantic relations are in the general case nonsymmetric, any pair of nodes can be linked by two edges with opposite directions (e.g., *finger* and *hand* could be linked by an edge directed from the former to the latter representing *part-whole*, and by an edge directed from the latter to the former representing the converse relation *whole-part*). Depending on the reasoner's knowledge about the analogs (as provided by textual information about whether and how particular concepts are related to one another), any concept pair in a graph can be connected by zero, one (unidirectional), or two (bidirectional) edges. In the simulations reported here, bidirectional links are constructed by default when the analogs consist of a simple set of words (as in the example depicted in Figure 1).

For more complex analogs presented as short texts, we explore the use of NLP techniques to identify keywords (words used frequently in the text, which correspond to core concepts) to serve as nodes in each semantic relation graph. For present purposes, we assume only very basic parsing of surface syntax to constrain generation of links. Words that appear close together in a text are more likely to be related in some significant way. As a simple heuristic to limit the size of semantic relation networks, we form links only between those keywords that cooccur within the same sentence. Unidirectional links are used to capture the directionality of subject-verb-object (noun-verb-noun) expressions. For example, the semantic relation network representing the sentence *dog chases cat* forms a triangle structure with three unidirectional edges to capture the head-to-tail pairwise relations (*dog →chase*, *chase → cat*, *dog → cat*). This directionality constraint can be applied to any noun-verb-noun structure in a text. We use this directionality constraint for all simulations with text input in the present paper.

Though doubtless oversimplified, these heuristics provide a preliminary procedure for using syntactic information conveyed by natural language to guide the construction of semantic relation networks (though some human intervention is still required). The general approach we favor is to extract as much guidance as possible from the surface syntax of text (without necessarily requiring the generation of more abstract propositional representations). We discuss NLP-assisted generation of semantic relation networks in connection with Simulation 5.

## Analogical Mapping with PAM

*Experiment and Simulation 1: Solving Analogies Based on Multiple Pairwise Relations*

PAM is able to resolve mapping ambiguities because the model maps concepts across analogs based on *patterns* of similarity among multiple pairwise relations. Patterns of similarity depend on semantic similarities between individual concepts (node attributes) and on relation similarities (edge attributes). We first examined mapping performance in humans and models for analogy problems in which resolving a mapping ambiguity requires integrating multiple relations in each analog: finding mappings between triplets of concepts that form an *A:B:C* category ordering, e.g., *weapon* : *gun* : *rifle* and *mammal* : *dog* : *beagle* (see the example in Figure 1). We created 12 Category triplets, and used all possible pairs as analogy problems to run a human experiment with a large sample size.

*Participants.* 1,329 participants ($M_{age}$ = 40.40, $SD_{age}$ = 11.98, age range = [18,82]; 711 female, 608 male, 6 gender non-binary, 4 gender withheld; minimum education level of high-school graduation, located in the United States, United Kingdom, Ireland, South Africa, or New Zealand) were recruited using Amazon Mechanical Turk (MTurk; approved, including informed consent procedures, by the UCLA Office of the Human Research Protection Program). Of these, 49 participants reported being distracted while completing the task, and their data were therefore excluded from analyses.

*Materials and Procedure.* Each participant completed two triplet analogy problems, one of each of two types. For each problem participants were asked to create a valid analogy by using their mouse to drag each of a set of randomly ordered terms (e.g., *mammal*, *beagle*, and *dog*) to one of the terms in an ordered set (e.g., *clothing : sweater : turtleneck*) presented in a fixed position on the screen (see Figure 3A). One problem was constructed out of 2 triplets in any order drawn from a pool of 12 (132 possible triplet pairs), each instantiating an *A:B:C* category ordering as in the example above (Category triplets). The other problem was constructed out of 2 triplets drawn from a different pool of 12 (another 132 possible triplet pairs), each instantiating a *part-whole* relation conjoined with an *object-location* relation (Part-Object-Location triplets). For example, a participant might be asked to match each of *fin*, *fish*, and *ocean* to each of *engine : car : garage*. The order of the two problems was counterbalanced across participants, and the specific triplets forming each problem were randomly assigned to each participant. By presenting each participant with just one problem of each type, we minimized any opportunity to learn the general structure of the problems (as our focus was on immediate analogical transfer, rather than schema induction). All triplets are provided in Supplemental Information, Table S1.

Before solving the two experimental problems, the analogy task was explained using two separate examples (involving different relations than the experimental problems). The instructions specified that an analogy is valid if the relations among the terms in each set match each other. Figure 3A depicts an example trial of the triplet mapping task as it would be performed by a human participant.

*Triplet Simulations.* We ran model simulations for all 132 analogy problems based on Category triplets. We compared the PAM model to several control models by varying relation representations and mapping algorithms. The relation vector was defined either as the concatenation of role and relation vectors created by BART (see Supplemental Information for additional variants), or as the difference of the Word2vec vectors for two words (Word2vec-diff), a standard procedure for forming a generic representation of the semantic relation instantiated by a pair of words (Zhila et al., 2013). The mapping algorithm was either the PAM model based on probabilistic mapping, or an alternative procedure based on exhaustive search of all possible mappings to maximize relation similarity. The exhaustive search algorithm represents the structure of a given triplet *A:B:C* as a concatenation of three vectors representing the pairwise relations between individual terms [*A:B*, *B:C*, *A:C*], and mapped an unordered triplet *D*, *E*, *F* to an ordered triplet *A:B:C* by finding the concatenated vector for the unordered triplet that yielded the highest cosine similarity to that of the concatenated vector for *A:B:C*. Note that the exhaustive search algorithm is only practical for relatively small analogy problems such as these triplet analogies, as the number of possible mappings is of the order $O(n!)$. For example, if an analogy problem involves 10 concepts in each analog, an exhaustive search of possible mappings would involve 3.6 million possible orders. In contrast, PAM based on probabilistic graduated assignment is much more efficient, with space complexity of $O(n^2)$. For simulations with PAM, a parameter search for $\alpha$ in Equation 4 found that a value of 0.1 yielded the highest mapping accuracy, indicating that accurate mapping in the triplet task requires downweighing the contribution of entity-based lexical similarity.

*Comparing Human and Model Performance.* For each analogy problem with Category triplets, the mapping response was counted as correct in an all-or-none manner: a mapping response was coded as 1 only if all three words were mapped correctly in a problem. Humans achieved mapping performance with average accuracy of 0.74. As shown in Figure 3B, the PAM model (PAM mapping algorithm coupled with BART) yielded the highest mapping accuracy of the four alternative models (0.83), exceeding mean performance of our MTurk participants. The two models using Word2vec-diff relation vectors were clearly inadequate. As reported in Supplemental Information, Table S2, standard PAM shows superior performance to further control models tested in ablation simulations.

Perhaps surprisingly, the PAM model yielded higher accuracy than the exhaustive search model with BART relation vectors (0.69). The exhaustive search algorithm considers all pairwise relations equally. However, due to the use of the balanced graph matching algorithm (Cour et al., 2006), PAM selectively emphasizes those edges that are more distinctive in their similarity pattern. For example, if the relation between nodes $i$ and $j$ in one analog shows high similarity to relations between many paired entities in the other analog, then its relation similarity is not discriminative in signaling the best mapping. In contrast, if the relation between nodes $i$ and $j$ in one analog shows high similarity only to the relation between nodes $i'$ and $j'$ in the other analog, and low similarity to relations between other paired entities, then its relation similarity is more informative in encouraging PAM to map *i* to *i'* and *j* to *j'*. In general, informative relations have a greater impact on mapping in PAM than do less informative relations, whereas the exhaustive algorithm weights all pairwise relations equally. Thus PAM is not simply a computationally tractable approximation to the exhaustive search algorithm; rather, PAM can lead to superior mapping performance for some analogy problems.

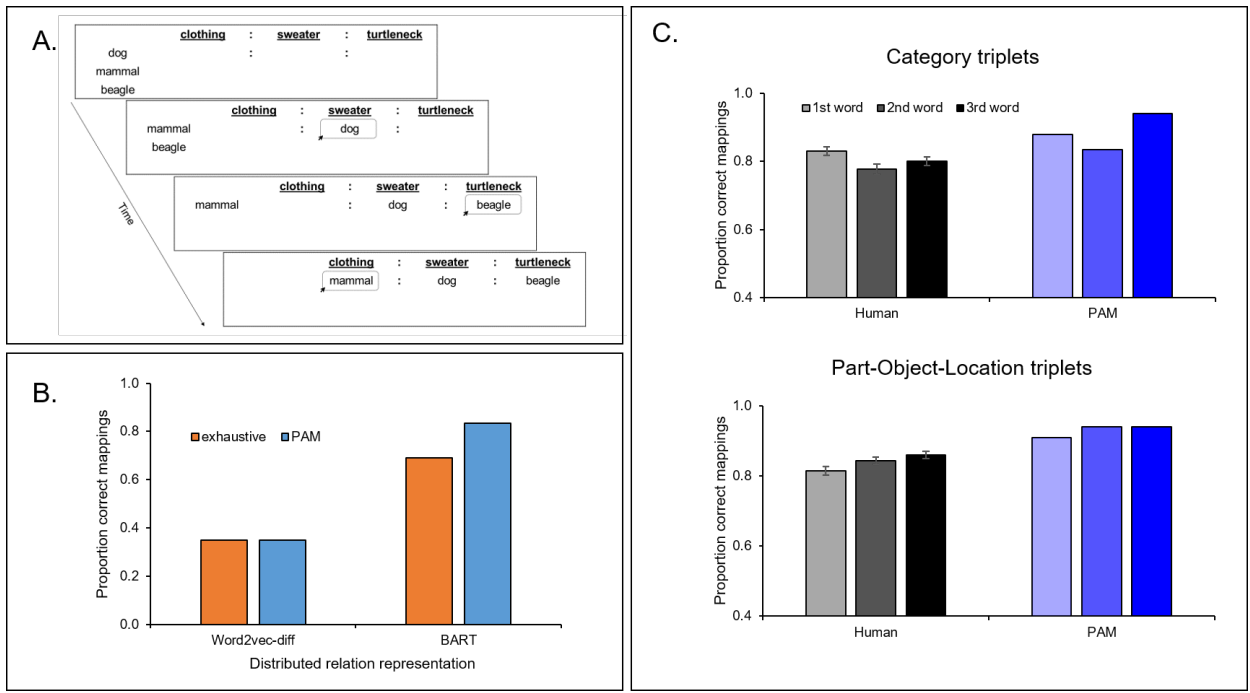

*Figure 3*. Model and human results for solving triplet analogies. (A) A mapping trial with two triplets of words. Participant must move each of three unordered words in one triplet (column at left) below the matching word in another triplet (top row) so as to form a valid analogy. (B) Simulation results (predicted proportion of triplet mappings entirely correct) for Category triplets using four alternative models: coding relations by vectors derived from Word2vec (difference vectors) or from BART, with mapping performed either by an exhaustive comparison of concatenated relation vectors or by the PAM algorithm. Human performance is indicated by the dashed line. (C) Mapping accuracy for words in the three positions within a triplet for humans and for PAM with BART vectors, separately for Category and Part-Object-Location triplets. Each datapoint for human performance is based on mean accuracy across individual participants who completed a given problem. Error bars indicate ±1 standard error of human accuracy for each word position across individual problems.

For both problem types, PAM's proportion correct was modestly higher than that observed for the human data (0.83 versus 0.74 for Category triplets; 0.89 versus 0.79 for Part-Object-Location triplets), perhaps due to variable effort on the part of the MTurk participants. In order to quantitatively compare predictions of PAM and the model variants to human performance, we conducted item-level analyses. We calculated the root-mean-square deviation (RMSD) of model predictions from the corresponding human responses (all responses scored as 1 for a fully correct mapping, .5 for a partially correct mapping, and 0 for an incorrect mapping) on each individual problem spanning both Category and Part-Object-Location triplets. Lower RMSD values indicate a closer match to human performance. The value of RMSD was lower for standard PAM (.22) than for either a control model without relation representations (Nodes-only, .38) or a control model with weak relation representations (Word2vec-diff with either mapping algorithm, .39). (For additional model comparisons see Supplemental Information.)

PAM also makes novel predictions regarding the accuracy of mappings at the level of individual word positions, which are predicted to vary across the two triplet types that were tested. As shown in Figure 3C, for Category triplets PAM predicts lowest accuracy for the middle word position (the intermediate category), whereas for Part-Object-Location triplets PAM predicts lowest accuracy for the first word position (the part). (See Supplemental Information for additional model comparisons.) Humans show a similar pattern of mapping accuracy at the level of word position in the different analogy problems. A two-way mixed ANOVA for mean human accuracy across problems, using triplet type (Category versus Part-Object-Location) as a between-problem factor, and word position (first vs. middle vs. last) as a within-problem factor, revealed reliable main effects for triplet type, $F(1, 262) = 6.24$, $p = .013$, and word position, $F(2, 524) = 3.53$, $p = .030$, as well as a reliable interaction, $F(2, 524) = 17.78$, $p < .001$. We followed up with six planned pairwise comparisons between word positions within each triplet type. Using a Bonferroni correction for multiple comparisons, we found that for Category triplets, participants were reliably less accurate in correctly mapping the middle word position (the intermediate category) than the first word position (the superordinate category), $t(131) = 4.48$, $p < .001$. The accuracy difference between the middle and last word positions (the subordinate category) fell short of significance after Bonferroni correction, $t(131) = 2.10$, $p = .207$, as did that between the first and last word positions, $t(131) = 2.21$, $p = .162$. For Part-Object-Location triplets, participants were least accurate for the first word position (the part): comparing first to middle word position (the object), $t(131) = 3.93$; comparing first to last word position (the location), $t(131) = 5.01$, both $p$'s $< .001$. Accuracy did not differ between the middle and last word positions, $t(131) = 1.57$, $p = .529$.

Overall, this set of findings confirms that PAM coupled with BART relation vectors is able to find systematic mappings by inferring and then integrating multiple pairwise relations, yielding mapping performance comparable to that of humans.

*Simulation 2: Mapping Science Analogies and Metaphors*

Analogies play an important role in both the development of scientific theories (Holyoak & Thagard, 1995) and in interpreting everyday metaphors (Lakoff & Johnson, 2003). It has generally been assumed that mapping such complex systems of knowledge depends directly on propositional representations, often using higher-order relations that take entire propositions as arguments. However, a study by Turney (2008) showed that people are able to find reasonable mappings for a set of 20 science analogies and analogical metaphors (Table 2) in which each analog has been reduced to 5-9 concepts corresponding to keywords, without accompanying texts (see Supplemental Information). These problems are all cross-domain, semantically-distant analogies, such as the Rutherford-Bohr analogy for the atom, or that between a computer and a mind. The keywords are a mix of nouns and verbs. For example, the source *solar system* is paired with the target *atom*, each represented by seven keywords (*solar system* set: *solar system*, *sun*, *planet*, *mass*, *attracts*, *revolves*, and *gravity*; *atom* set: *atom*, *nucleus*, *electron*, *charge*, *attracts*, *revolves*, and *electromagnetism*). To take a second example, the source analog *computer* includes 9 keywords (*computer*, *outputs*, *inputs*, *bug*, *processing*, *erasing*, *write*, *read*, *memory*), as does the target analog *mind* (*mind*, *muscles*, *senses*, *mistake*, *thinking*, *forgetting*, *memorize*, *remember*, *memory*). Turney showed that this set of analogies can be solved reliably by a computational model (Latent Relation Mapping Engine) that searches large text corpora for relation words associated with each keyword, and then uses frequencies of co-occurrence as an index of relational similarity. Turney also asked 22 human participants to map words between source and target analogs, and assessed human performance in this mapping task for each problem.

*Table 2.* Science analogies and analogical metaphors dataset (20 source/target pairs) developed by Turney (2008). Number of keywords for each problem appears in parentheses.

| *Science analogies* | *Analogical metaphors* |
| --- | --- |
| A1. solar system / atom (7) | M1. war / argument (7) |
| A2. water flow / heat transfer (8) | M2. buying an item / accepting a belief (7) |
| A3. water waves / sound waves (8) | M3. grounds for a building / reasons for a theory (6) |
| A4. combustion / respiration (8) | M4. physical travel / problem solving (7) |
| A5. sound waves / light waves (7) | M5. money / time (6) |
| A6. terrestrial motion / planetary motion (7) | M6. seeds / ideas (7) |
| A7. agricultural breeding / natural selection (7) | M7. machine / mind (7) |
| A8. billiard balls / heat due to molecular motion (8) | M8. holding object / understanding idea (5) |
| A9. computer / mind (9) | M9. path following / argument understanding (8) |
| A10. slot machine / bacterial mutation (5) | M10. seeing / understanding (6) |

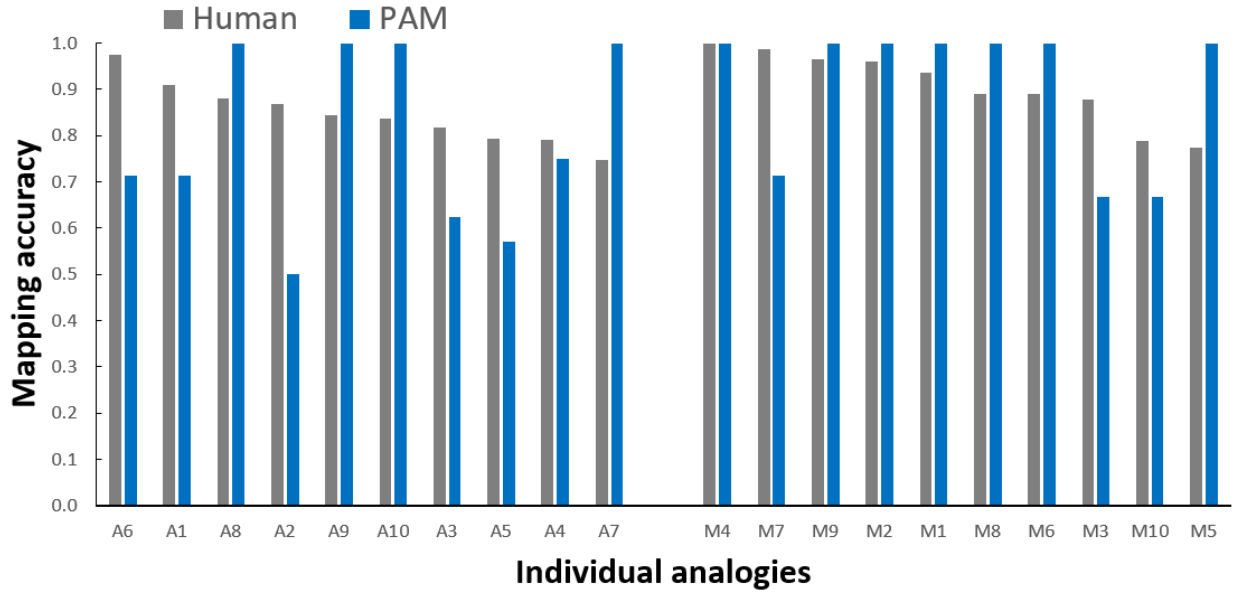

*Figure 4.* Model and human performance for solving 10 science analogies and 10 analogical metaphors in a dataset developed by Turney (2008). Numbers correspond to analogies listed in Table 2 (ordered from highest to lowest human accuracy within each type). Each problem includes 5-9 keywords in source and target analogs (numbers are provided in Table 2). Mapping accuracy is defined as the proportion of keywords correctly mapped.

We applied PAM to this dataset of 10 science analogy problems and 10 analogical metaphors. In Simulation 2, PAM formed semantic relation networks for each analog, with node attributes coded as Word2vec vectors for each word, and edge attributes coded as BART vectors for all pairwise relations. It is important to note that none of the word pairs in these materials had been used to train the BART model. This simulation thus provides a strong test of generalization for relation identification, based on the distributed representations of relations created by BART.

Figure 4 depicts human and PAM mapping accuracy for each problem. Across the 20 analogies, the PAM model achieved a mean accuracy of 85% in identifying correct correspondences of keywords between two analogs, approaching the 88% accuracy observed in Turney's (2008) experiment with adult human participants (though less than the 92% accuracy achieved by Turney's LRME model). Trends for individual problems showed higher mapping accuracy for PAM over humans for 10 problems, humans over PAM for 9, with one tie. For the computer/mind analogy, PAM yielded correct mappings between all corresponding words across the two analogs (*computer* to *mind*, *outputs* to *muscles*, *inputs* to *senses*, *bug* to *mistake*, etc.).

Several model variants were also tested (see Supplemental Information). Given that the analogs used in Simulation 2 are based solely on keywords, without any support from structured text, it is reasonable to consider whether performance of the full PAM model was driven solely by the semantics of the keywords themselves. In general, PAM's performance was fairly similar across model variants. However, overall accuracy was reduced if edge similarity was excluded (Nodes-only, 77%), or if Word2vec-diff was used instead of BART to create relation vectors (Word2vec-diff, 77%). Hence PAM's performance depended in part on BART's relation vectors. The standard PAM model correctly predicts lower mapping accuracy for the 10 science analogy problems (0.79) than for the 10 analogical metaphor problems (0.90), consistent with the difference observed for humans (0.85 for science analogy problems and 0.91 for analogical metaphor problems). The model variant without relations (Nodes-only) did not show this difference (.78 accuracy for science analogy problems and .77 for analogical metaphor problems). Accuracy remained high (0.85) for the BART-role variant, in which the edge vectors included only the role component. Recall that for each relation in BART's vector, the role component is jointly determined by the relation and by semantic features of the keyword playing the first role of the relation. Thus although relation information played a significant role in PAM's performance, semantic features of keywords were certainly influential.

*Simulation 3: Pragmatic Influences on Mapping*

Some analogies pose mapping ambiguities that cannot be resolved simply by integrating the available relations, because the relations themselves support multiple potential mappings about equally. For example, when people were asked to draw analogies between the actors in the first

Gulf War (in 1991) and those in World War II, the American President George H. W. Bush paired with United States was sometimes mapped to Franklin Roosevelt and the United States, and sometimes to Winston Churchill and Great Britain (as all three pairs instantiate the relation "wartime leader of nation"; Spellman & Holyoak, 1992). Individual participants tended to choose one or the other of the two mappings that were pairwise consistent, with a significant number choosing each. Pairs were almost always mapped consistently (i.e., people seldom mapped Bush to Roosevelt but the United States to Great Britain).

Although people clearly prefer isomorphic (one-to-one) mappings, they must cope with naturalistic situations of this sort that have considerable relational overlap (as well as similarities between individual objects), but that are not in fact isomorphic. In such cases people sometimes give responses that violate a strict one-to-one constraint (e.g., about 7% of participants mapped the United States of the Gulf War era onto both the United States of World War II and also Great Britain). In several experiments using non-isomorphic analogies, a minority of participants produced one-to-many or (more often) many-to-one mappings (Spellman & Holyoak, 1992, 1996; Krawczyk, Holyoak, & Hummel, 2004). Such findings are consistent with mapping models such as PAM (also ACME, Holyoak & Thagard, 1989, and LISA, Hummel & Holyoak, 1997) that treat isomorphism as a soft constraint, rather than a strict filter on possible mappings.

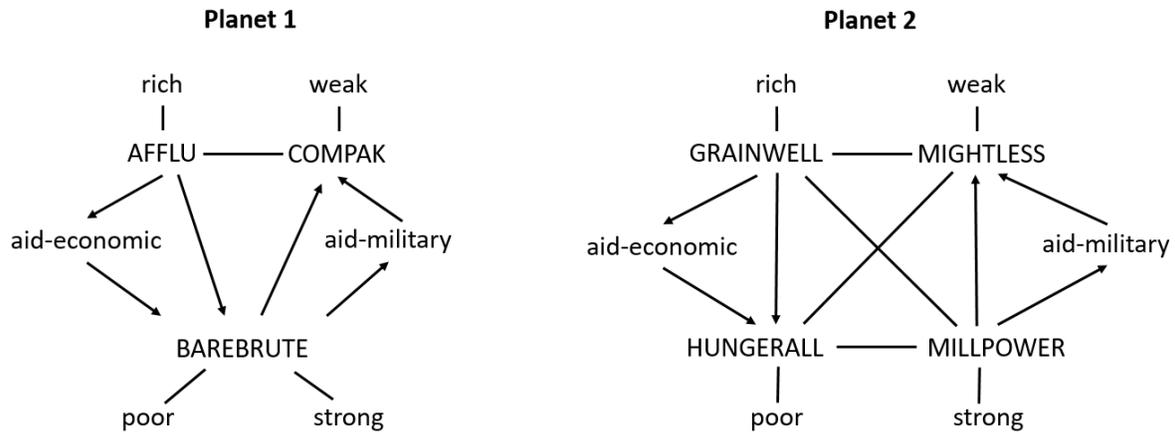

*Figure 5.* Semantic relation networks for story analogs from Spellman and Holyoak (1996, Experiment 2).

Particularly when analogs are in fact non-isomorphic, a preferred mapping may be determined by pragmatic factors, notably the reasoner's goal in using the analogy (Holyoak, 1985). In PAM, prior beliefs about probable analogical correspondences, and preferences for correspondences based on goal-related elements, can be represented by varying attention weights (Nosofsky, 1986) on relevant nodes and edges (reflecting relative attention to different components of analogs). In Simulation 3, PAM was applied to a set of non-isomorphic story analogies used in a study by Spellman and Holyoak (1996, Experiment 2). Each analog was a science-fiction-style description of multiple countries. The countries on each of two planets (forming the source and target analogs) were linked by various economic and/or military alliances, such that the country Barebrute on one planet could be mapped to either the country Hungerall on the second planet based on a shared economic relation (summarized by the predicate *aid-economic*), or to Millpower based on a shared military relation (*aid-military*; see schematic description in Figure 5). In these stories, similarities were balanced so that the mapping for Barebrute was ambiguous, as country Barebrute on Planet 1 had equal similarity to the countries Hungerall and Millpower on Planet 2.

In the human experiment, manipulations of participants' processing goals guided their preferred mappings: stressing the importance of either economic factors or Hungerall encouraged the Barebrute—>Hungerall mapping relative to the Barebrute—>Millpower mapping, whereas stressing either military factors or Millpower encouraged the Barebrute—>Millpower mapping relative to the Barebrute—>Hungerall mapping.

PAM was provided with a set of concepts for each analog. As shown in Figure 5, the source analog included 9 concepts (*Afflu, Barebrute, Compak, rich, poor, strong, weak, aid-economic, aid-military*), and the target analog included 10 concepts (*Grainwell, Hungerall, Millpower, Mightless, rich, poor, strong, weak, aid-economic, aid-military*). Note that the predicates *aid-economic* and *aid-military* were included as nodes. The Word2vec vector for *country* was assigned to all the imaginary countries used in both analogs (hence node similarity could not discriminate among the possible mappings for any country). To simulate the pragmatic impact of goals on mapping the ambiguous country (Barebrute), attention weights were increased on the relations relevant to a particular goal. In the condition stressing the importance of Hungerall, all pairwise relations involving Hungerall, including its relation to Grainwell via *aid-economic*, were assigned attention weights. In the condition stressing economic factors, the relation between Grainwell and *rich* was also emphasized. Complementary sets of relations were assigned attention weights to model the conditions emphasizing the country Millpower or military factors.

During each simulation run, PAM sampled the value of its attention weight from a uniform distribution within the range of [1, 1.1]. The left panel in Figure 6 shows the proportion of trials on which humans selected Hungerall or else Millpower as the preferred mapping for the ambiguous country Barebrute across different conditions. The right panel in Figure 6 shows the probability that PAM selected each preferred mapping (obtained by averaging 1000 samples of attention weights for each experimental condition). In the simulation of the control condition (equal emphasis), PAM predicts that the ambiguous country Barebrute will be mapped to Hungerall and Millpower with equal probability. Although PAM does not include an explicit decision mechanism for dealing with such ambiguous mappings, it would be reasonable to expect that a reasoner would either provide both correspondences or report just one of the two (chosen randomly). In Spellman and Holyoak's (1996) experiment, about half of human participants in the control condition mapped both Hungerall and Millpower to Barebrute, so many-to-one mappings are certainly observed.

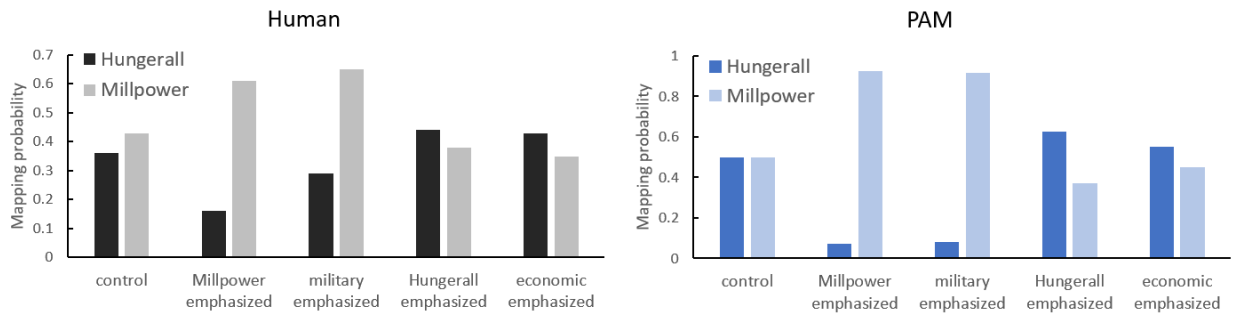

*Figure 6.* Human responses (Spellman & Holyoak, 1996, Experiment 2) and PAM predictions in Simulation 3, based on the probability of mapping the ambiguous country Barebrute to either Hungerall (based on similar economic relations) or Millpower (based on similar military relations) for different experimental conditions.

When attention weights are used to emphasize particular goals, PAM predicts the qualitative shift in the preferred mapping for Barebrute. Given emphasis on Hungerall or economic factors, Barebrute is more likely mapped to Hungerall; given emphasis on Millpower or military factors, Barebrute is more likely mapped to Millpower. It is also noteworthy that PAM captures an asymmetry in the impact of goals on mapping. Both in the human data and in PAM's predictions, the impact of emphasis on Hungerall or economic factors yields a smaller shift in mappings relative to an emphasis on Millpower or military factors. This asymmetry emerges even though PAM models the economic and military goals using exactly symmetrical shifts in attention weights.

This asymmetry arises from subtle differences in the semantic similarities among keywords (nodes), which are inherited from the Word2vec embeddings that encodes semantic meanings of individual concepts: *economic* shows more similar semantic associations to several alternative keywords (*poor*, *rich*, *weak*, *strong*) than does *military*. These semantic differences render the military-related keywords more distinctive (resulting in less ambiguous mappings), and hence more resistant to displacement when competing with economic-based mappings than vice versa. When model variants were examined, this asymmetry disappeared in the variant that excluded node similarities (BART edges-only), confirming that keywords similarities drove the effect (see Supplemental Information). Edge similarities were certainly critical to the overall performance of the model, as the variant that omitted edge similarities (Nodes-only) was completely unable to capture the pattern of human mapping judgments. The Word2vec-diff model was unable to run at all because of issues arising from the identical node vectors (for the word *country*) assigned to all the imaginary countries. Only the full PAM model, which fully integrates rich semantics of concepts and relations into the mapping process, can account for the fine-grained aspects of human judgments.

*Simulation 4: Modeling the Relational Shift in Cognitive Development*

Developmental research has shown that children's ability to reason and solve problems by analogy generally improves with age (Holyoak, Junn, & Billman, 1984). Over the course of cognitive development, children undergo a *relational shift* (Gentner & Rattermann, 1991) from a primary focus on direct similarity of objects towards greater reliance on relational information. As noted previously, a variety of factors are known to globally shift the balance between the impact of semantic similarity of individual concepts versus relational similarity on comparison judgments. Relational similarity tends to be more potent when overall relational similarity across analogs is relatively high (Goldstone et al., 1991), when the objects in visual analogs are sparse rather than rich (Gentner & Rattermann, 1991; Markman & Gentner, 1993), and when participants are given more time to make their judgments (Goldstone & Medin, 1994). From the perspective of PAM such global factors, including the relational shift, can potentially be captured by variations in the parameter $\alpha$ in Equation 6.

As a test of whether PAM can account for developmental changes in analogy performance, we simulated findings from a classic developmental study by Gentner and Toupin (1986). In this study children used toys to enact interactions among three animal characters, using actions familiar to young children (e.g., playing). The experimenter guided the children to act out the events in the source story, and then asked them to repeat the same events using different characters in a target analog. Using variants of the same basic stories, a 2x2 design was created to manipulate *systematicity* of the source analog and *compatibility*[1] of the mappings between source and target characters. In one example, the source text was a short passage describing interactions among a seal, penguin, and dog. The introduction to the systematic version stated that the seal was jealous and did not want his friend the penguin to play with anyone else. The story ended by stating the moral that it is wrong to be jealous and better to have more friends. In the nonsystematic version the introduction described the seal as "strong" rather than "jealous", and the moral was omitted. The body of the story was identical in the two versions, describing how the penguin played with the dog, angering the seal, but how the dog eventually saved the seal from danger. Thus in the systematic version the seal's jealousy had an intuitive connection to his anger and behavior, which the nonsystematic version lacked. Gentner and Toupin (1986) hypothesized that the systematic version was richer in higher-order relations (in particular, the relation *cause*, presumed to hold between propositions).

The target analog also involved three animal characters. In the high compatibility condition three animals, each similar to one of the source characters, played corresponding roles: <*seal*, *penguin*, *dog*> maps to <*walrus*, *seagull*, *cat*>. High compatibility between relational roles and object similarity makes mapping straightforward. In the low compatibility condition the same animals were assigned to different roles, creating cross mappings between roles and object similarity: <*seal*, *penguin*, *dog*> maps to <*seagull*, *cat*, *walrus*>. Such cross mappings, in which the tendency to match relational roles competes with a preference to match similar objects, create difficulty for children and even for adults (Markman & Gentner, 1993). Finally, in the medium compatibility condition the three animals to be mapped were <*lion*, *giraffe*, *camel*>, which have no particular similarity to animals in the other analog. The medium compatibility condition was thus neutral with respect to mappings between relational roles and animals.

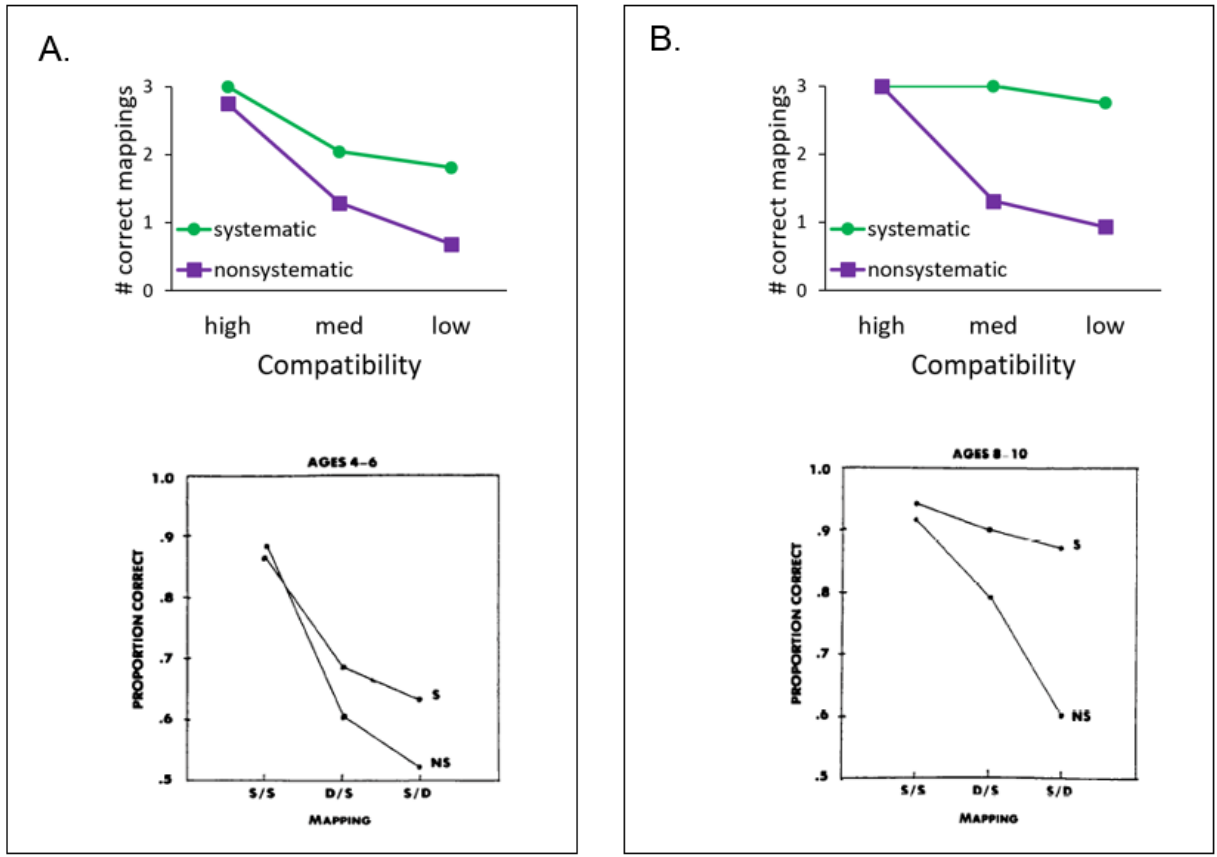

*Figure 7*. PAM simulation of character mapping by children of different ages (top) and children's accuracy in reenacting stories (bottom; reprinted by permission from Gentner & Toupin, 1986). A, top: Mapping accuracy with node attributes weighted with $= 2$ ; bottom, data for 4-6 year old children. B, top: Mapping accuracy with node attributes weighted with $\alpha = 1$; bottom, data for 8-10 year old children.

Gentner and Toupin (1986) tested groups of younger (4-6 years) and older (8-10 years) children, scoring each child on accuracy in reenacting the story using the target characters. Children in both groups performed best when the source was systematic and the character mapping was highly compatible. Accuracy was impaired when the source was nonsystematic and when compatibility was low. The overall advantage of the systematic analogs was attributed to the presence of higher-order (causal) relations that participate in the mapping. For the younger children, systematicity was less beneficial overall than it was for the older children, and the detrimental impact of cross mappings (low compatibility) was greater. These findings were interpreted in terms of a relational shift, such that the older children were less influenced by object similarity and guided more by relational information (especially higher-order relations).

The PAM simulation provides a qualitative approximation to the experiment with children (using mapping accuracy as an estimate of accuracy in story reenactment). Simulations were based on eight different sets of animal triplets used in the experiment (see Supplemental Information for materials and also additional model comparisons). The keyword concepts used as input to PAM were drawn from the introduction to the stories: three animals and the verb *play*. These keywords were shared across the systematic and unsystematic conditions. The only difference between the inputs for the two versions was that the systematic version included the adjective *jealous* as a keyword, whereas the substituted adjective *strong* was not included as a keyword for the unsystematic version. *Jealous* was selected as a keyword because it occurs twice in the systematic version (both in the introduction and the concluding moral), whereas *strong* occurs only once in the unsystematic version. This selection was confirmed by running two different NLP algorithms on the stories (a program developed by Tixier, Skianis, & Vazirgiannis, 2016, and the MATLAB TextRank keyword extraction function). Both algorithms identified *jealous* but not *strong* as a keyword. More generally, we hypothesize that high systematicity reflects greater text coherence (Kintsch, 1988), for which keyword extraction provides a simple proxy. Notably, PAM was not provided with any information about higher-order relations for either the systematic or unsystematic versions.

To distinguish the syntactic subject and object roles in noun-verb-noun sentences, PAM uses unidirectional edges. In this simulation, given a sentence stating that one animal plays with another (e.g., *seal plays penguin*), the forward edge only was created for the three constituent pairwise relations (*seal* : *play*, *play* : *penguin*, *seal* : *penguin*). These single forward edges capture the directionality of the syntactic subject and object roles associated with the verb. For example, given the source analog *seal plays penguin*, PAM will give a higher mapping score for the potential target analog *walrus plays seagull* than for the cross-mapped source analog *seagull plays walrus*, yielding a positive effect of compatibility.

We manipulated the parameter $\alpha$ in Equation 6, which controls the impact of lexical similarity (nodes) in mapping. PAM compared mapping performance when lexical similarity was given a high weight (Figure 7A, top, $\alpha = 2$) to performance when this weight was lower (Figure 7B, top, $\alpha = 1$). Greater weight on node attributes increases the impact on analogical mapping of similarity between animals in the source and target relative to the impact of semantic relations (edge attributes), simulating the lesser sensitivity to relation similarity for preschool children. As shown in Figure 7, the PAM simulation captures important trends in development of ability in analogical reasoning: the overall benefit of both source systematicity and mapping compatibility, the interaction between these two factors, as well as greater sensitivity to systematicity for older children. PAM's simulation demonstrates that the impact of systematicity on analogical mapping, at least in the study by Gentner and Toupin (1986), can be explained without requiring any assumptions about the use of higher-order relations.

With respect to the relational shift, theorists have suggested that the developmental increase in focus on relations may reflect some mix of increased relational knowledge and maturation of important executive processes, notably working memory and inhibitory control (Morrison, Doumas, & Richland, 2011; Richland, Morrison, & Holyoak, 2006). To model changes in reasoning abilities of the 4-6 year versus 8-10 year old children, PAM holds constant both knowledge of relations (with no higher-order relations being used at either age) and the basic mapping algorithm. Instead, PAM simply manipulates its parameter $\alpha$, controlling the relative influence of relation versus object similarity on mapping. PAM thus instantiates the hypothesis that with increasing age, children come to preferentially place greater weight on relations when reasoning by analogy. This account is consistent with more recent proposals about the development of analogical reasoning, which attribute some performance differences among children to learned (possibly culturally-dependent) preferences for relations versus objects, rather than differences in either prior knowledge or processing ability (Richland, Chan, Morrison, & Au, 2010; Kuwabara & Smith, 2012; Carstensen, Zhang, Heyman, Fu, Lee, & Walker, 2019; also see Kroupin & Carey, 2021). More generally, the relational shift is very likely the product of multiple types of developmental changes.

*Simulation 5: Mapping Richer Text Representations*

PAM can potentially map more complex analogies presented as texts. As an initial effort, we applied the model to find mappings between a source story and target problem introduced by Gick and Holyoak (1980) and widely used in psychological research on analogical problem solving. The source story (*The General*) describes how a general sends small groups of soldiers down multiple roads to capture a fortress located in the center of a country; the target (radiation problem) describes a doctor attempting to use a kind of ray to destroy an inoperable stomach tumor without damaging healthy tissue. The analogous convergence solution to the radiation problem is to use multiple weak rays directed at the tumor (Duncker, 1945). The two analogs are not isomorphic, and not all concepts have clear mappings.

The texts for the two analogs (Supplemental Information, Table S7) consisted of 13 sentences constituting the problem statement for *The General* and 7 sentences constituting the radiation problem. Keywords and semantic relations were identified using an NLP-assisted procedure. After replacing pronouns with their antecedents, each text was passed through a customized text pre-processing code with the following four steps: (1) the top 20 most frequent words were selected; (2) the MATLAB TextRank keyword extraction function was run to select nouns, adjectives, and verbs included in the top 20 high-frequency word set; (3) the MATLAB bag-of-n-grams model with the window size of 2-gram was used to identify pairwise relations between the keywords; (4) noun-noun relations were selected only when both nouns appeared in the same sentence. The procedure for keyword extraction and pairwise relation pre-processing yielded 14 keywords for *The General* story (*country, dictator, fortress, villages, commander, large, army, capture, roads, landmines, attack, small, many, troops*) and 10 keywords for the radiation problem (*doctor, tumor, patient, destroyed, ray, high, intensity, destroy, healthy, tissue*). To avoid extreme heteronyms, the word *general* was replaced by *commander* and *mines* were replaced by *landmines*. Given their near synonymy, *troops* was replaced by *army* in forming pairwise relations. We then manually added an important *noun-verb-noun* relation to each representation, "commander controls army" and "doctor uses rays". Although not directly stated in the texts, both are obvious inferences. As in previous simulations, unidirectional edges were used to code *noun-verb-noun* relations. The NLP-assisted text pre-processing thus yielded the basic elements for representing the analogs. Semantic relation networks were created for each analog, with Word2vec embeddings providing node attributes for individual keywords, and BART vectors providing edge attributes for semantic relations between keywords.

Figure 8 shows the major correspondences between concepts that PAM identified for the two analogs. Seven concepts have mappings for which humans generally agree: *army→ray*, *fortress→tumor*, *commander→doctor*, *country→patient*, *large→high*, *capture→destroy, controls→uses*). PAM identifies all seven of these major mappings, and also maps *attack* to *destroy*. According to human intuition, a few concepts in *The General* have no clear match in the radiation problem (e.g., dictator, landmines, roads). Because PAM aims to find mappings for all concepts, these poorly-matched concepts end up in arbitrary pairings.

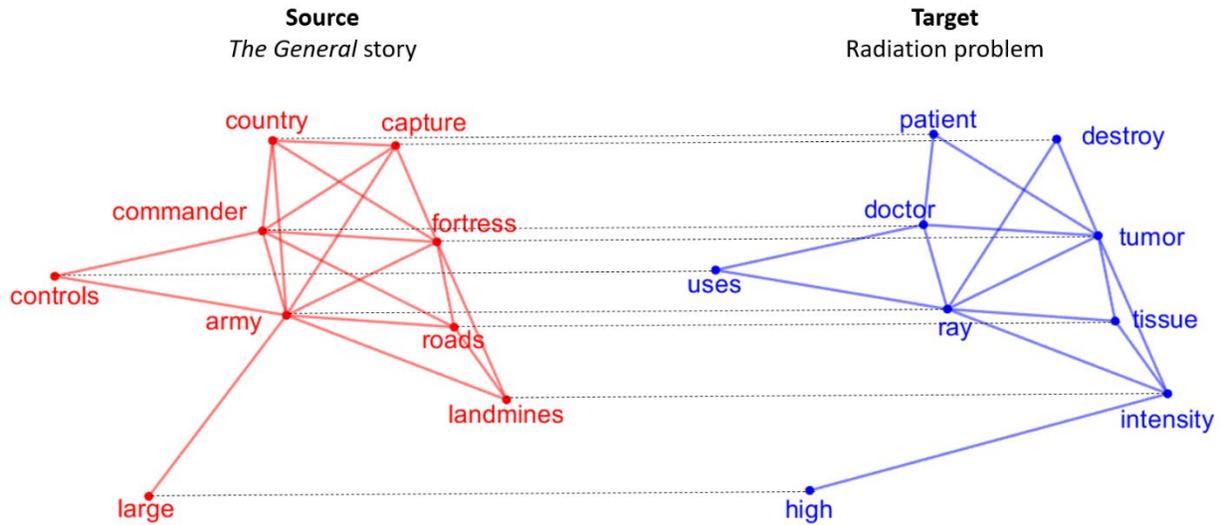

*Figure 8.* Semantic relation networks for *The General* story and the radiation problem, with mappings created by PAM that link major concepts. (To avoid clutter some keywords are omitted.)

We performed additional simulations using variant models (see Supplemental Information). In particular, we sought to confirm that PAM's mapping performance is not solely due to the structure of its semantic relation networks, but also depends on the content of the relation vectors. When relation vectors produced by BART were replaced with Word2vec-diff vectors (while holding constant all other operations in the mapping algorithm), only two of the seven major correspondences between concepts were recovered. Accuracy was also reduced when the full PAM model was ablated to include only node similarities (Nodes-only, four correct correspondences) or only edge attributes (BART edges-only, three correct). Thus PAM's performance critically depends on having effective representations of semantic relations as edge attributes, and not solely on the form of its semantic relation networks.

*Simulations 6 and 7: Selection of Plausible Source Analogs*

As noted earlier, mapping is considered central to analogical reasoning because it impacts other key processes, including the initial retrieval of a potentially useful source analog. Memory retrieval involves a process of comparing a cue to cases stored in long-term memory; although it seems unlikely that a cue could be mapped to every stored case, several models of analog retrieval have proposed that some form of mapping can be performed on a smaller set of "finalists" that pass some basic threshold of similarity (Forbus et al., 1995; Hummel & Holyoak, 1997; Thagard et al., 1990).

Indeed, analog retrieval appears to be sensitive to the same basic constraints as analogical mapping: similarity of individual concepts and of relations, modulated by differential attention to goal-relevant elements. However, retrieval and mapping are at least partially dissociable. Gick and Holyoak (1980, 1983) showed that people often fail to retrieve a potentially useful far analog, suggesting retrieval is more difficult than mapping when the analogs are semantically distant. Later studies confirmed this pattern (Gentner, Rattermann, & Forbus, 1993; Holyoak & Koh, 1987; Keane, 1987; Ross, 1987, 1989; Seifert, McKoon, Abelson, & Ratcliffe, 1986). Other studies found that people produce far source analogs in response to a target more often when the pragmatic context involves a goal of communicating ideas to others (e.g., political argumentation; Blanchette & Dunbar, 2000, 2001). However, in the latter studies the relative number of near and far potential analogs actually available in memory was unknown. When relative availability is taken into account, studies using naturalistic paradigms also find that analog retrieval is heavily influenced by similarity of individual concepts (Trench & Minervino, 2015). At the same time, experimental evidence indicates that relational similarity also impacts retrieval (Wharton et al., 1994; Wharton, Holyoak, & Lange, 1996).

Analog retrieval, of course, is simply one manifestation of the normal operation of memory processes. The problem of finding a "useful" source analog in response to a target analog can be considered as a special case of a rational analysis of memory (Anderson & Milson, 1989). Given a set of potential sources $S$ stored in memory, the probability that $S_i$ is the optimal candidate for retrieval given some target[2] analog $T$ will be proportional to the prior probability that $S_i$ is optimal multiplied by the likelihood that $S_i$ is optimal given $T$. The prior will favor source analogs that have been useful in the past; hence highly familiar sources will tend to be preferentially retrieved. A particularly well-established example is the prevalent use by children of the "person" concept as a source in making inferences about other animals and plants (Inagaki & Hatano, 1987). Similarly, people understand new individuals by spontaneously relating them to significant others, such as a parent or close friend (Andersen, Classman, Chen, & Cole, 1995).

Analogical access, like memory retrieval in general, is inherently competitive. Studies have shown that for any cue, people are more likely to retrieve a case from long-term memory if it is the best match available (based on similarity of both concepts and relations) than if some other stored case provides a better match (Wharton et al., 1994, 1996). Retrieval involves competition between multiple potential source analogs, whereas mapping focuses on a single source. Also, during mapping information about both analogs can be actively processed in working memory, enabling eduction of relations to create or elaborate semantic relation networks. In contrast, source analogs are necessarily dormant in long-term memory prior to being retrieved.

In terms of PAM, encoding a source analog would involve storage of all or part of a semantic relation graph. Storage of concepts and relations will be inherently asymmetrical: concepts $A$ and $B$ (nodes in a graph) may be stored without necessarily storing the relation $A{:}B$, whereas $A{:}B$ (the edge between $A$ and $B$) can only be stored if $A$ and $B$ are also stored. Since retrieval of far analogs will be relatively dependent on shared relations, retrieval of far versus near analogs will be disadvantaged to the extent important semantic relations in the source were not fully encoded into memory. As this analysis predicts, domain experts (who are more likely to focus on relevant relations during both encoding and retrieval) are more successful than novices in accessing remote source analogs based on relational similarity (Novick, 1988; Novick & Holyoak, 1991; Goldwater, Gentner, LaDue, & Libarkin, 2021).

Given that semantic relation networks have been stored in long-term memory, PAM can be used to calculate a measure of overall similarity—a "mapping score", which is the log-likelihood defined in Equation 4 using the maximum a posteriori estimate of mapping $\widehat{\boldsymbol{m}}$ inferred by the model. We term this measure the $G$ score:

$$G = \sum_i \sum_{j \neq i} \sum_{i'} \sum_{j' \neq i'} \widehat{m}_{ii'} \widehat{m}_{jj'} S(A_{ij}, A_{i'j'}) + \alpha \sum_i \sum_{i'} \widehat{m}_{ii'} S(A_{ii}, A_{i'i'}). \qquad (7)$$

$G$ provides an overall assessment of similarity based on both concept similarity (node attributes) and relational similarity (edge attributes). This mapping score can be used to rank alternative source analogs in degree of fit to a target.[3] We report two simulations to illustrate the use of mapping scores to predict analog retrieval.

Simulation 6 aimed to demonstrate that PAM can account for the often-observed dissociation between the impact of concept similarity on retrieval versus mapping. Keane (1987, Experiment 1) examined retrieval of several variations of source analogs to the radiation problem (see Simulation 5). The source analog was always presented as a story, which was studied 1-3 days before presentation of the target radiation problem. Keane found that 88% of participants retrieved a source analog from the same domain as the target (a story about a surgeon treating a brain tumor), whereas only 12% retrieved a source from a remote domain (a story about a general capturing a fortress, very similar to that used by Gick & Holyoak, 1980). This difference in ease of access was dissociable from the ease of post-access mapping and inference: the frequency of generating the convergence solution to the radiation problem once the source analog was cued was high and equal (about 86%) regardless of whether the source analog was from the same or a different domain.

The source texts used by Keane (1987) were very similar to those examined in Simulation 5, but shorter and simpler. For the military story (far source analog) our text pre-processing program yielded six keywords (*commander*, *fortress*, *army*, *country*, *destroy*, *large*); for the medical story (near analog) the program yielded seven keywords (*surgeon*, *cancer*, *rays*, *brain*, *destroy*, *high*, *intensity*). The radiation problem (target) yielded 10 keywords. We ran PAM the same way as reported in Simulation 5. PAM found the correct mapping of core concepts for both source stories: *commander* → *doctor*, *fortress* → *tumor* and *army* → *rays* for the far source; and *surgeon* → *doctor*, *cancer* → *tumor* and *rays* → *rays* for the near source. However, the value of the $G$ score was substantially higher for the near source (2.216) than for the far source (1.665). Simulation 6 thus captures the qualitative pattern of human performance: higher probability of retrieving a near than far source, coupled with equal (and high) success in mapping either to the target.

In Simulation 7 we examined PAM's predictions for analog retrieval using a larger dataset. Turney's (2008) dataset of 20 science analogies and analogical metaphors (used in Simulation 2) provides a test of PAM's ability to identify "good" source analogs for a given target analog. The mapping score $G$ can be used to rank order any number of potential source analogs in degree of fit to a given target. To assess the ability of PAM to rank potential source analogs, in Simulation 7 the 20 examples of analogies in Table 2 were used to create a source selection task. For this simulation, we assume that the semantic relation networks for all 20 source analogs have been successfully stored in long-term memory, and that the semantic relation for a given target analog serves as a retrieval cue. For example, given *atom* (#1) as the target analog, we would expect *solar system* to be selected as the best source analog among the 20 alternatives.

For this computational experiment, each target analog in turn was mapped to all 20 source analogs, and PAM's mapping score was obtained for each potential source. For 17 out of the 20 targets, the best match as computed by PAM was the intended source shown in Table 2. For the three target analogs in which PAM's mapping score did not identify the expected source analog, the "errors" proved to be very reasonable. The science analogy dataset includes target analogs that are closely linked to multiple relevant sources. Specifically, two of the science analogies involved gravity (#1 and #6 in Table 1), two involved wave motion (#3 and #5), and two involved heat (#4 and #8). PAM assessed the best match for *planetary motion* (#6) to be *solar system* (#1), a source analog based on essentially the same knowledge (with overlapping keywords). The best match for the target *sound waves* (#3) was the source *sound waves* (#5, with somewhat different keywords), and the best match for *heat due to molecular motion* (#8) was the source *combustion* (#4). These choices indicate that PAM favors retrieval of near over far analogs, as do humans. In each of these cases PAM selected the intended source shown in Table 2 (a far analog) as the next-best match. For the target *light waves* (#5) the expected source (*sound waves*) was selected as the best match, with the other available wave analog (*water waves*, #3) ranked second. Figure 9 shows the entire distribution of *G* scores across the 20 alternative source analogs for each of four representative target analogs.

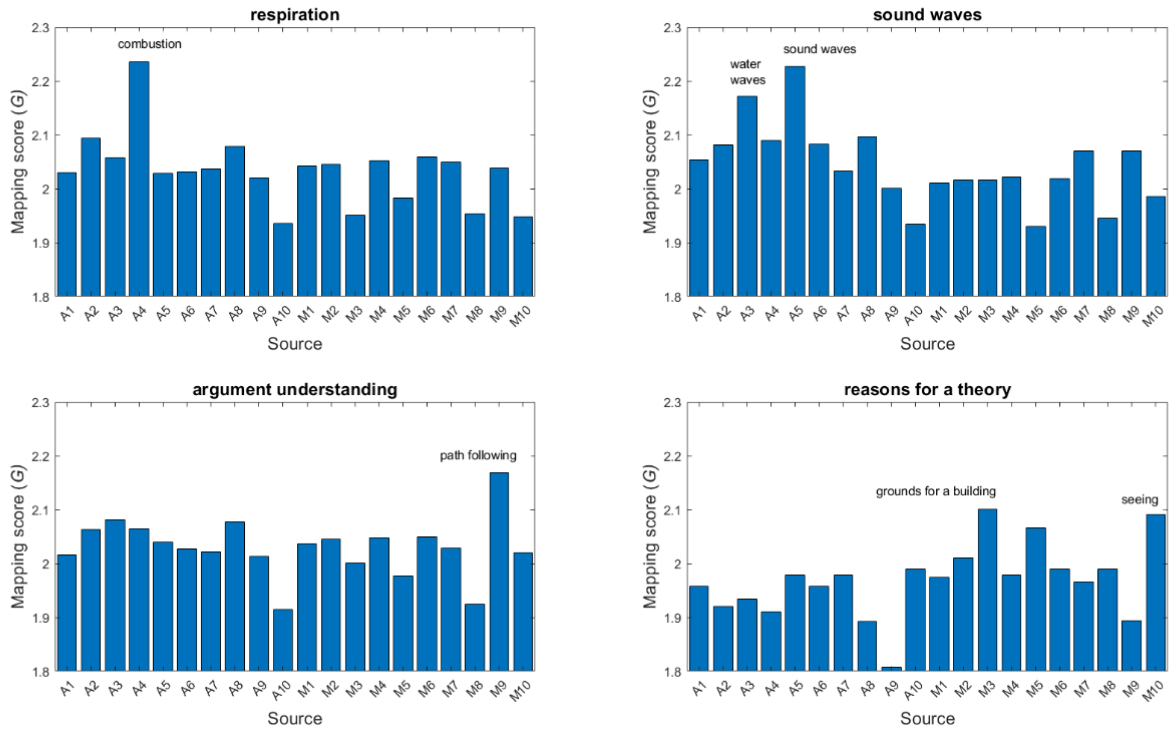

*Figure 9.* Representative examples, for each of four target analogs, of the distribution of *G* scores across 20 alternative source analogs. Each panel shows scores for one target in Turney's (2008) dataset; the horizontal axis indicates 20 possible source analogs (labels identified in Table 2). A: *respiration* correctly and unambiguously maps to *combustion*; B: *sound waves* map to *sound waves* (very near analog) closely followed by *water waves* (intended far analog); C: *argument following* correctly and unambiguously maps to *path following*; D: *reasons for a theory* maps correctly to *grounds for a building*, followed closely by two alternative source analogs.

For each target analog, PAM's measure of mapping quality thus identified relationally-similar source analogs from semantically-distant domains. The results of Simulation 7 indicate that when a semantic relation network has been stored in long-term memory for each analog in a set of potential source analogs, PAM can select a useful source for a given target from among a larger pool of candidates. Of course, if the semantic relation networks for source analogs were poorly encoded or have been degraded by forgetting, retrieval will be less sensitive to shared relations.

## General Discussion

*Summary and Implications*

We have presented a novel Bayesian model of analogical mapping, PAM, which builds on advances in machine learning that automate the generation of rich semantics for both concepts (by Word2vec) and relations (by BART trained using Word2vec). From these inputs, the model creates semantic relation networks that capture the skeletal structure of complex analogs. PAM operates on semantic relation networks to find mappings between key concepts. The model automatically creates relation networks by integrating fragmentary semantic knowledge about individual concepts and the relations that link them, guided by textual constraints when available.

PAM is able to solve analogies that require integration of multiple relations in each analog (Simulation 1, based on data from a novel mapping task), matching the pattern of human performance in considerable detail. The model is also able to solve complex analogical mappings based on sets of predefined keywords (Simulation 2, based on Turney, 2008), and can account for the impact of goals in resolving ambiguous mappings (Simulation 3, based on Spellman & Holyoak, 1996). By varying the model's global emphasis on lexical concept (node) versus relation (edge) similarity, it is possible to account for developmental shift in sensitivity to relations; in addition, the model can capture the influence of text coherence (systematicity) without assuming access to higher-order propositions (Simulation 4, based on Gentner & Toupin, 1986). By adding NLP-assisted pre-processing to extract key concepts from text inputs, the model can solve analogies between non-isomorphic problems posed in short texts (Simulation 5, based on Gick & Holyoak, 1980). The model also provides a measure of global similarity between analogs, which can be applied to support the retrieval of plausible source analogs from memory, and accounts for the partial dissociation between the impact of different types of similarity on retrieval versus mapping (Simulation 6, based on Keane, 1987; Simulation 7, based on Turney, 2008).

The use of semantic relation networks in analogical mapping is consistent with empirical evidence that human memory and comprehension are reconstructive in nature, and heavily dependent on prior semantic knowledge (Kintsch, 1988; Van Overschelde & Healy, 2001). PAM captures the human ability to reason by analogy in a domain-general manner, without requiring extensive training with analogy problems in any particular domain. The model's success in accounting for a range of phenomena observed in studies of human analogical mapping, as well as analog retrieval, illustrates how distributed representations capturing rich semantics of concepts and relations can effectively accomplish tasks that are usually associated with symbolic reasoning (Carstensen & Frank, 2021; Holyoak & Lu, 2021).

PAM represents the continuing evolution over the past four decades of models of human analogical mapping based on representation matching. An early hypothesis was that mapping is based solely on structure and not on content (Gentner, 1983); however, empirical evidence led subsequent computational models to add constraints based on semantic content and pragmatic factors (e.g., Gentner et al., 1993; Forbus et al., 2017; Holyoak, 1985; Holyoak & Thagard, 1989). In PAM, not only individual concepts (nodes in a semantic relation graph) but also semantic relations themselves (edges) are defined by rich semantic vectors. As demonstrated in comparisons between the performance of PAM using relation vectors generated by BART versus Word2vec-diff, mapping performance can differ radically depending on semantic content even when the form (structure) of the semantic relation networks being compared is held constant (see model comparisons in Supplemental Information, especially for Simulations 1 and 5). In general, for the simulations reported here, PAM approaches human levels of performance only when it operates on relation vectors provided by BART. The present computational results are in accord with previous evidence that BART, but not Word2vec-diff, generates semantic relation vectors that can account for human patterns of relational similarity judgments (Ichien et al., 2021) as well as patterns of neural similarity among relations observed during analogical reasoning (Chiang et al., 2021). At the same time, Word2vec proved very effective in providing semantic vectors for individual concepts (nodes), which are also crucial for PAM's performance. Perhaps most notably, subtle variations in patterns of node similarity predicted a previously-unexplained asymmetry in resolution of an ambiguous mapping, observed in a study by Spellman and Holyoak (1996; Simulation 3). Thus semantic content—of both concepts and relations—is fundamental to human analogical mapping.

Like previous models in the tradition of representation matching, PAM is able to capture the human ability to perform zero-shot learning by analogical transfer. In the spirit of other recent work in this tradition (Doumas et al., 2020; Forbus et al., 2017), the model makes progress (though still incomplete) toward achieving a key aim emphasized by end-to-end models: automating the generation of analog representations. PAM, coupled with Word2vec and BART, enables semi-automated generation of the inputs to the mapping module for analogies presented in verbal form, either as sets of keywords or as short texts. When the inputs are texts, the initial extraction of key concepts and their relations is aided by NLP techniques that serve as proxies (albeit imperfect) for human text comprehension. Given the limitations of the NLP techniques we have so far explored, this process requires some human intervention. But once the nodes and edges in semantic relation graphs have been specified, the remainder of the mapping process is fully automated. By building a mapping model on top of learning mechanisms grounded in distributional semantics, we can draw closer to the goal of being able to automate analogical reasoning for natural-language inputs.

*Forms of Relational Representation*

The BART/PAM framework for learning relations and reasoning with them exemplifies the general view that (at least for humans) relational knowledge in acquired by a series of *re-representations* (Penn, Holyoak, & Povinelli, 2008). These successive representations of relational knowledge lie on a continuum from highly implicit to increasingly explicit (see Doumas & Hummel, 2012). Within the model presented in the present paper, four distinct representations that include relational information can be distinguished. (1) Word embeddings created by Word2vec take the form of densely distributed vectors representing the meanings of individual concepts (words). Within these embeddings, some features carry relational information, but in an implicit and "entangled" manner (Moradshahi et al., 2021). To take a hypothetical example, the embeddings for the words *rich* and *poor* may each include features associated (in a probabilistic manner) with such relational concepts as money, continuous quantity, and relative extremity. (2) Given word pairs that form positive and negative examples of a target relation, the learning mechanisms in BART take advantage of statistical information coded by Word2vec features (coupled with derived features created by reordering based on the magnitudes of feature differences between the words in a pair) to generate a weight distribution over word features, thereby predicting the posterior probability of the target relation (e.g., *opposite*). This weight distribution for a relation provides a new implicit representation that captures the relative importance of various word features in predicting the relation. (3) The computed posterior probability that the target relation holds is then treated as a representation of the degree to which a word pair satisfies the relation, coded as the value of a disentangled "relation feature" in a new representational space of relations. (4) BART forms an explicit semantic-relation vector composed of values (i.e., posterior probabilities) of the features corresponding to its learned relations. As a disentangled and explicit (though distributed) representation, this vector makes it possible to form PAM graphs in which the specific relation between two words (edge attribute) is distinct from the meanings of the individual keywords being related (node attributes). Importantly, these successive relation representations do not replace one another, but rather support complementary cognitive abilities. Most notably, the weight distributions learned by BART accomplish the eduction of relations for any word pair, whereas the resulting semantic vectors allow analogical comparisons (the eduction of correlates).

Semantic relation vectors do not exhaust the forms in which relational knowledge can be represented. These vectors are distinct from verbs and other linking words that constitute multi-place predicates in natural language—a yet more explicit form of relation representation. The relation vector for a particular pair of words need not correspond to a predicate; however, specific relation features (individually or as a set) may be linked to corresponding predicates (e.g., *contrast* features might be connected to the phrase "is opposite of", and *part-whole* features to "is a part of"). At the same time, most verbs and other multi-place predicates (e.g., *chase*, *kill*, *love*, *give*) probably do not directly correspond to features in a semantic relation vector. Philosophers have distinguished between relations that are *internal*—those that hold by nature of the terms they relate—versus *external* (e.g., Clementz, 2014). Word embeddings tend to capture *generic* information about concepts (Cimpian & Markman, 2008; Graham, Gelman, & Clarke, 2016)— properties that are intrinsic to the meaning of the words (*pansy* is a type of *flower*) or highly typical (*read* is an action performed on a *book*). Given that the semantic relations learned by BART are extracted from generic feature representations of their relata (i.e., word embeddings), these relations can be construed as internal. Semantic relation vectors thus focus on internal relations between generic word meanings, whereas predicates of a natural language can refer to all types of relations. Thus a model such as BART, which aims to learn semantic relation vectors, has a different (though ultimately related) goal than does a model such as DORA (Doumas et al., 2008), which focuses on learning predicates.

*Coding Sentential Information in Semantic Relation Graphs*

PAM operates on attributed graphs consisting of nodes representing individual concepts and edges representing the educed relations between concepts (Spearman, 1923), rather than on structured propositions as assumed by previous models of analogical mapping in the tradition of representation matching. An important issue concerns whether and how semantic relation networks can capture the detailed structural information provided by sentences of natural language (or by abstract propositions derived from sentences). An obvious limitation of semantic relation vectors is that all relations are binary (connecting two words), whereas the predicates of natural language also include (at least) ternary relations (*b is between a and c*; *x gave y to z*). In our simulations of analogies expressed by text, we introduced a convention for coding subject-verb-object sentences as sets of unidirectional edges in a graph (e.g., "dog chases cat" becomes the trio of binary relations *dog : chase*, *chase : cat*, *dog : cat*). This convention captures at least part of the relational structure conveyed by simple sentences, enabling PAM to predict the greater difficulty of cross mappings (see Simulation 4 based on Gentner & Toupin, 1986). The basic approach of translating simple sentences into a set of unidirectional binary links could be extended to sentences that include an indirect object (e.g., "The boy gave a book to the girl"), which express ternary relations. By adopting NLP techniques that create relatively "flat" syntactic parses (perhaps with a version of dependency grammar; see Jurafsky & Martin, 2021, Chapter 14), the general approach could be further extended to handle sentential complements (a type of sentence embedding), such as "The woman believed that the boy loved the girl." Moreover, a parser could be used to augment semantic relation vectors with features indicating syntactic (e.g., subject versus object) and/or thematic roles (e.g., agent versus patient).

It remains an open question whether analogical mapping requires sensitivity to aspects of relations that cannot be fully captured by semantic relation networks. It has often been claimed that analogical mapping (at least for older children and adults) depends on representations of "higher-order" relations (those that take propositions as arguments, typically corresponding to verbs that take sentential complements). Higher-order relations have been assumed to increase the "systematicity" of mappings (Gentner, 1983). However, Simulation 4 (Gentner & Toupin, 1986) raises the possibility that at least some evidence for the impact of systematicity can be explained without positing access to higher-order relations at all. An alternative general explanation is that mapping benefits from greater text coherence (Kintsch, 1988), which does not necessarily depend on higher-order propositions. Within the PAM model, mapping performance will generally be facilitated by any and all textual cues that establish unambiguous semantic relations between multiple pairs of interconnected keywords.

In fact, the majority of studies that have been interpreted as supporting the special importance of higher-order relations in mapping solely involve the relation *cause*, which has been treated as a higher-order relation between propositions describing events (e.g., Clement & Gentner, 1991; Gentner et al., 1993; Forbus et al., 2017). However, linguistic evidence casts doubt on this representational assumption. Verbs that express direct causation commonly appear in single-clause sentences (e.g., "The boy broke the vase"), rather than taking sentential complements (Kemmer & Verhagen, 1994; Wolff, 2003). Moreover, other relations that appear syntactically equivalent to *cause* (e.g., *temporally prior to*) do not support analogical inferences in the same manner (Lassaline, 1996). Indeed, causal relations have long been viewed as having a special status in analogical reasoning—not because of their syntactic form, but because of their pragmatic relevance to goal attainment and hence analogical inference (Winston, 1980; Holyoak, 1985). Theories of causal reasoning have treated causal relations not as static predicates attached to higher-order propositions, but rather as active components of causal networks that generate inferences (Pearl, 2009; Holyoak & Cheng, 2011; Waldmann & Hagmayer, 2013). Analogical inferences are sensitive to such basic distinctions as that between generative and preventive causes, and predictive (cause to effect) versus diagnostic (effect to cause) inferences (Holyoak, Lee, & Lu, 2010; Lee & Holyoak, 2008). Thus although causal relations indeed have special properties, their impact on analogical reasoning may have little to do with the syntactic form of causal propositions.

*Limitations and Future Directions*

Whatever role (direct or indirect) that propositions may play in analogical mapping, such representations are likely to be important for other aspects of analogical reasoning. In particular, PAM has yet to be extended to address the later stages, inference and schema induction. While semantic relation networks enable flexible and computationally-efficient analogical mapping, more detailed propositional representations (or at least the syntax of natural language) may well be required to enable construction of explanations for mappings, and to aid in the generation of structured analogical inferences. The immediate output of PAM is simply a set of mappings between individual keywords from the source and target. However, if the analogs were presented as structured text, the basic algorithm of "copy with substitution" (CWS; Holyoak, Novick, & Melz, 1994) can be used to generate analogical inferences. For example, the mapping between the *The General* story and the radiation problem (Simulation 5) yields correspondences that include

*commander*→*doctor* and *army*→*ray*. If the source contains the sentence, "The commander divides the army", CWS would yield the inference "The doctor divides the ray"—a valuable aid in constructing a parallel convergence solution to the radiation problem. More generally, the two types of explicit relation representations (semantic relations and predicate-centered propositions) may prove to be complementary. PAM can construct semantic relation networks and use them to produce a quick sketch of the mapping between two analogs, coupled with an evaluation of overall mapping quality. If the mapping appears to be promising, propositional representations can potentially be used to develop the mapping in greater detail and to generate inferences from it.

A number of extensions of the PAM model appear feasible. The relation vectors provided by BART could be improved by training the model on a broader range of relations, particularly the types of thematic relations that link verbs with nouns. Mapping could in principle be based in part on embeddings of concepts derived from larger units than words, including sentences (Devlin et al., 2019), and visula embeddings derived from images via convolutional neural networks. PAM's procedure for iterative updating of the mapping matrix could include a slack column of allowing concepts that do not map well to go unmatched. The iterative search for an optimal mapping might allow the value of the α parameter to vary (i.e., seeking a trade-off between a focus on node versus edge similarity that maximizes mapping quality) using hierarchical Bayesian modeling approach.  A further extension might allow the search for an optimal mapping to include variations in the encodings of the analogs, as suggested by the Copycat model (Hofstadter & Mitchell, 1994). Such extensions are likely to lead to a greater emphasis on sequential processing, which is likely necessitated by capacity constraints on human analogical reasoning (Halford et al., 1998; Hummel & Holyoak, 1997; Keane & Brayshaw, 1988).

Although the present paper deals only with verbal analogies, PAM can be used to perform mapping given any system for assigning vectors as attributes of nodes and edges in a graph. The model could therefore be adapted to solve mappings based on perceptual inputs such as pictures, given that relevant object features and perceptual relations have been identified. Relation vectors for meaningful pictures can potentially be formed as hybrids of features provided by perceptual processes and by semantic knowledge about concepts (Lu, Liu, Ichien, Yuille, & Holyoak, 2019). It seems possible that even formal relations, such as those used in psychological studies of the acquisition of relational schemas (Halford, Bain, Maybery, & Andrews, 1998; Halford & Busby, 2007; Phillips, 2021), as well as those that occur in mathematics, can also be represented by relation vectors with semantic content. Indeed, studies such as that by Halford et al. (1998) have shown that learning of new formal relations can be facilitated by encouraging participants to map them onto known meaningful relations. Similarly, people's interpretation and use of arithmetic operations appears to be guided by *semantic alignment* between mathematical and real-life situations. The entities in a problem situation evoke semantic relations (e.g., tulips and vases evoke the functionally asymmetric *contain* relation), which people align with analogous mathematical relations (e.g., the non-commutative *division* operation: tulips/vases; Bassok & Olseth, 1995; Bassok, Chase, & Martin, 1998). A similar form of semantic alignment guides the use of different formats for rational numbers—fractions and decimals. Adults in the United States and South Korea (Lee, DeWolf, Bassok, & Holyoak, 2016), as well as Russia (Tyumeneva et al., 2018), selectively use fractions and decimals to model discrete (i.e., countable) and continuous entities, respectively. Favored semantic alignments may reflect selective similarities between relation vectors that represent mathematical and real-world relations.

PAM may be able to contribute to efforts to automate the discovery of analogies in online databases (e.g., by searching an inventory of patents for inventions). A processing pipeline might use algorithms for natural language processing to summarize texts stored in electronic form and to extract key concepts coupled with basic syntax. The extracted information would then be processed to form knowledge graphs with rich semantics for concepts and relations, which could in turn be passed to PAM to identify potential source analogs relevant to solving a specified target problem. Our broader aim is to foster the evolving synergy between theoretical ideas drawn from AI and from cognitive science, in order both to understand human reasoning more fully and to enhance the reasoning capacities of machines.

## Acknowledgements

Preparation of this paper was supported by NSF Grant IIS-1956441 to H.L. and NSF Grant BCS-1827374 to K.J.H. We thank Alex Doumas, John Hummel and Taylor Webb for helpful discussions, and four anonymous reviewers for valuable comments on an earlier draft. The anonymized data for the triplet experiment are available on Open Science Framework. Reviewers can access the data using this view-only link: https://osf.io/u3wmz/?view_only=0871683f4a3a4984835480d3dbd3f564. Reviewers can also access the Qualtrics code that generates the triplet experiment using this view-only link: https://osf.io/u3wmz/?view_only=0871683f4a3a4984835480d3dbd3f564 All code for the PAM model will be made available on GitHub. A version of the manuscript has been posted on arXiv: https://arxiv.org/abs/2103.16704

## Footnotes

1. The variable we refer to as "compatibility" (of object and relational similarity) was termed "transparency" by Gentner and Toupin (1986). For methodological reasons, in their experiment the characters were varied in the source rather than target, which creates the same 2x2 design used in our simulation.

2. Here we encounter an unfortunate conflict between the target/source terminology of the analogy literature and the cue/target terminology of the memory literature. In typical analog retrieval, a target analog serves as a retrieval cue, and potential source analogs are available as "targets" stored in long-term memory. We will continue to use the term "target" in the sense of a target analog.

3. Equation 7 can potentially be refined to deal with situations in which semantic relation graphs for different analogs vary substantially in size (cf. Marshall, 1995). In addition, a quantitative model would relativize retrieval probability to reflect competition among alternative source analogs stored in memory (Hummel & Holyoak, 1997). In the present simulations the analogs are of similar size, and we only make qualitative predictions (aiming to predict the rank order of retrieval probabilities for different possible source analogs).

# Supplementary Information

**Probabilistic Analogical Reasoning with Semantic Relation Networks**

Hongjing Lu        Nicholas Ichien        Keith J. Holyoak

Correspondence to: Hongjing Lu (hongjing@ucla.edu)

**This file includes:**

## I. Materials for Triplet Analogy Experiment

*Table S1.* Materials used to construct triplet analogy problems.

| **Category triplets**<br><br>*Superordinate : Intermediate : Subordinate* | **Part-Object-Location triplets**<br><br>*Part : Object : Location* |
|---|---|
| dairy : cheese: cheddar | fin : fish : ocean |
| vegetable : lettuce : romaine | wing : bird : nest |
| grain : rice : basmati | paw : bear : cave |
| fruit : melon : honeydew | stinger : bee : hive |
| reptile : lizard : iguana | petal : flower : garden |
| fish : salmon : sockeye | leaf : tree : forest |
| bird : parrot : parakeet | prickle : cactus : desert |
| mammal : dog : beagle | root : grass : meadow |
| weapon : gun : rifle | propeller : airplane : airport |
| clothing : sweater : turtleneck | engine : car : garage |
| vehicle : car : SUV | hull : boat : marina |
| tool : ax : hatchet | siren : ambulance : hospital |

## II. Simulation 1: Model Comparisons for Triplet Analogy Experiment

We compared PAM's performance in solving the triplet analogies to that of nine control models. The models differed in (a) how they represented constituent pairwise relations and (b) the mapping algorithm used. Four alternatives for forming relation vectors linking were pairs were assessed: (1) difference vectors computed directly from Word2vec vectors for head and tail keywords (Word2vec-diff); (2) the vector of role probabilities for the head word computed by BART, without relation information (BART-role); (3) the vector of relation probabilities computed by BART, without role information (BART-rel); (4) the concatenated role and relation vectors respectively used in BART-role and BART-rel (BART-full). In addition, for PAM control model ablation simulations, we tested a Nodes-only version that lacks relation representations altogether, and a BART edges-only version that includes full BART vectors (relation plus roles) without node vectors. In all model variants, all relations were coded as bidirectional. For PAM simulations, we used lexical similarity weight of $\alpha = 0.1$ and BART relation vectors with a nonlinear power transformation that emphasizes the strongest elements of the relation vector (power of 5, based on model comparisons reported by Ichien et al., 2021).

Two algorithms were compared. One was the PAM algorithm for graph matching based on graduated assignment. The other was an exhaustive algorithm in which the concatenation of the three pairwise relations in the source triplet (presented in a horizontal row with order fixed; see Figure 2A in main text) was compared to each of the 6 possible ordered concatenations of the vectors for the three pairwise relations in the target triplet (words presented in random order in vertical column, with relations *A:B*, *B:C* and *C:D*). For each of the 6 comparisons between concatenated vectors for source and target, cosine similarity was calculated, and the target vector yielding the highest similarity with the source vector was used to define the predicted mapping of words in the target to those in the source.

For all models, overall accuracy was calculated as the proportion of the 132 distinct mapping problems of each type for which the model generated the complete correct mapping. Human accuracy for complete problems was scored in the same way. In order to make more detailed comparisons between model and human performance at the level of individual problems, we calculated the root-mean-square deviation (RMSD) of each model's responses from the corresponding human responses on each individual problem spanning both Category and Part-Object-Location triplets. For the RMSD analysis, all responses (for models and humans) were scored as 1 for a fully correct mapping, .5 for a partially correct mapping, and 0 for an incorrect mapping.

Results are summarized in Table S2. The standard PAM model yielded the highest overall performance, achieving both the highest accuracy and the smallest deviation from human performance across all model variants. Notably, the Nodes-only model (no relations) results in a drop in overall accuracy and an increase in its deviation from human performance, compared to the standard PAM model. On the other hand, removing the effect of lexical similarity (BART edges-only model) does not harm performance much at all. This is not surprising, since the standard PAM model for this task uses a low weight for lexical similarity ($\alpha = 0.1$). This asymmetry in the impact of edges and nodes emphasizes the importance of relation information in analogical mapping, even for these simple 3-term mapping problems.

Moreover, both PAM and the Exhaustive mapping algorithm showed considerable sensitivity to the particular kind of relation representation used to construct its edges. Using Word2vec-diff relation representations with both algorithms consistently yields a stark drop in performance, compared to the various BART relation representations, demonstrating the importance of semantic relation content on analogical mapping.

*Table S2.* Comparison of standard PAM (PAM + BART-full) and control models with human accuracy on triplet analogy problems for Simulation 1. Root-mean-square deviation (RMSD) values indicate the average deviation of model responses on individual problems from human responses on corresponding problems (lower values indicate closer approximation to human performance). The RMSD is calculated for all questions for both types of triplets. Boldface indicates the best model performance, which is achieved by the standard PAM model.

| Mapping Algorithm | Relation representations | Mapping accuracy Category triplets | Mapping accuracy on Part-Object-Location triplets | Item-level analysis: RMSD from human performance |
|---|---|---|---|---|
| Exhaustive mapping | Word2vec-diff | .35 | .68 | .39 |
| | BART-role | .65 | .85 | .30 |
| | BART-rel | .71 | .89 | .28 |
| | BART-full | .69 | .86 | .29 |
| PAM | Word2vec-diff | .35 | .68 | .39 |
| | Nodes-only | .33 | .67 | .38 |
| | BART edges-only | .82 | .88 | .22 |
| | BART-role | .79 | .89 | .23 |
| | BART-rel | .65 | .89 | .30 |
| | **BART-full** | **.83** | **.89** | **.22** |
| Mean Human Accuracy | | .74 | .79 | |

In addition to examining overall performance on the analogy triplets, we also examined the extent that the control models could account for variability in mapping accuracy for each word position within Category and Part-Object-Location triplets. As described in the main text, the standard PAM model correctly predicts the lowest accuracy for the middle word position in Category triplets, and the lowest accuracy for the first word position in Part-Object-Location triplets. Figure S1 presents the proportion of correct mappings made by the PAM model and four variants of it (see Table S1). (The BART edges-only model is omitted from Figure S1, as its predictions are almost identical to those of standard PAM.) In general, the models that include variants of BART relations produce human-like performance (except that the BART-rel variant fails to match the human pattern of position effects for Category triplets). The models that lack any sort of BART relations (Nodes-only and Word2vec-diff) are considerably less accurate than humans, and fail to capture the pattern across word position for Part-Object-Location triplets. This discrepancy highlights the importance of BART's disentangled relation representations for PAM's ability to generate human-like mapping performance.

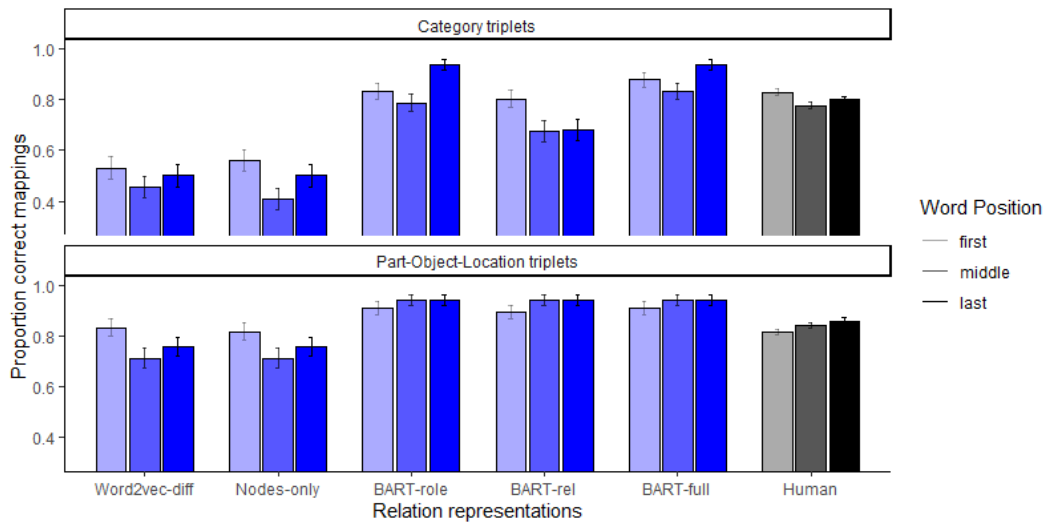

*Figure S1*. Proportion of correct mappings, broken down according to word position within analogy triplets, for model variants and humans. Top panel depicts performance on Category triplets; bottom panel depicts performance on Part-Object-Location triplets.

### III. Simulation 2: Model Comparisons for Mapping Science Analogies and Metaphors

Keyword representations of 10 science analogies and 10 common metaphor problems were adopted from Turney (2008). Each analogy involves a source domain and a structurally similar target domain between which a knowledgeable human reasoner can identify a rich set of relational correspondences. The keyword representations reduce each of these analogical domains to two sets of 5–9 keywords. For example, the source *solar system* is paired with the target *atom*, each represented by 7 keywords (*solar system* set: *solar system*, *sun*, *planet*, *mass*, *attracts*, *revolves*, and *gravity*; *atom* set: *atom*, *nucleus*, *electron*, *charge*, *attracts*, *revolves*, and *electromagnetism*). Near synonyms were substituted for multi-word synonyms (e.g., *activate* for *turn on*). Word2vec representations of keywords were used as the input to PAM. For PAM simulations, we used lexical similarity weight of $\alpha = 1$ and BART relation vectors with a nonlinear power transformation (power term = 5). All the model variants for PAM (see Table S2) were tested. As shown in Table S3, the best model performance was achieved by both the standard PAM model and the variant using BART role vectors only, suggesting that for this set of problems the role component of the relation representation was most important.

*Table S3.* Comparison of standard PAM (PAM + BART-full) and control models with human accuracy on Turney dataset for Simulation 2. Mapping accuracy is calculated by proportion of correctly identified correspondences of keywords between two analogs, and averaged across all 20 problems. Boldface indicates the best model performance, which is achieved by the standard PAM model and the variant with BART role vectors.

| Mapping Algorithm | Relation representation | Mapping accuracy |
|---|---|---|
| PAM | Word2vec-diff | .77 |
| | Nodes-only | .77 |
| | BART edges-only | .67 |
| | **BART-role** | **.85** |
| | BART-rel | .73 |
| | **BART-full** | **.85** |

PAM's performance on individual problems can be improved by assuming knowledge about basic noun-verb-noun structure. For example, in the solar system/atom analogy, PAM mapped 5 word pairs correctly, mistakenly mapping *solar system* to *electron* and *sun* to *atom*. However, if unidirectional edges are added for *noun-verb-noun* structures (e.g., *planet revolves sun* is coded by unidirectional edges for *planet* : *revolves*, *revolves* : *sun*, and *planet* : *sun*), then PAM is able to match all seven keywords correctly.

### IV. Simulation 3: Model Comparisons for Pragmatic Influences on Mapping

PAM was applied to a set of non-isomorphic story analogies used in a study by Spellman and Holyoak (1996, Experiment 2). Each analog was a science-fiction-style description of multiple countries. As shown in Figure 5 in the main paper, the source analog included 9 concepts (*Afflu, Barebrute, Compak, rich, poor, strong, weak, aid-economic, aid-military*), and the target analog included 10 concepts (*Grainwell, Hungerall, Millpower, Mightless, rich, poor, strong, weak, aid-economic, aid-military*). The same Word2vec embedding vector for *country* was assigned to all the imaginary countries used in both analogs (hence node similarity could not discriminate among the possible mappings for any country). For PAM simulations, we used lexical similarity weight of $\alpha = 1$ and BART relation vectors with nonlinear power transformation (power term = 5) for the simulation. During each simulation run, PAM sampled the value of its attention weight from a uniform distribution within the range of [1, 1.1]. The model results were computed as the mean mapping probability averaged across 1000 runs. All the model variants for PAM (see Table S2) were tested.

Given the complexity of the experimental design, we focused model comparisons on two key qualitative aspects of the human data. First, for each model, we determined whether it predicted the qualitative shift in the mapping for the ambiguous country (Barebrute) depending on the pragmatic focus on military or else economic relations. Second, we determined whether each model predicted the asymmetry in this shift (more pronounced for the military than for the economic conditions). Finally, we calculated the correlation between model predictions and human response proportions across the 10 conditions in the design (see Figure 6 in main paper for human data and predictions of the standard PAM model).

As summarized in Table S4, the highest correlation (.96) was achieved by the standard PAM model and the two variants that included BART roles and/or relations. These three models all were able to account for the two key qualitative differences. However, good model performance also required inclusion of node vectors: performance of the variant with BART edges-only was poor. The nodes-only model and the Word2vec-diff model failed completely. The former model, lacking any relations that could be influenced by attention weights, generated identical predictions for response proportions across all experimental conditions. The latter model failed to yield any solution. Because all the imaginary countries were coded by identical node vectors (the vector for *country*), they generated cosine distances of 0. In summary, BART vectors proved to be critical in accounting for the influence of pragmatic factors on human mapping.

*Table S4.* Comparison of PAM and control models with human performance for Simulation 3 of pragmatic impact of goals on mapping (Spellman & Holyoak, 1992, Experiment 2), based on predicted and observed mappings for the ambiguous country (Barebrute). Boldface indicates the best model performance, which is achieved by the standard PAM model and the variants based on BART roles and/or relations.

| Mapping Algorithm | Relation representations | predict the qualitative shift in mapping ambiguous country? | predict asymmetry in shift (Military vs. Economic)? | correlation with human mapping proportion |
|---|---|---|---|---|
| PAM | Word2vec-diff | no | no | undefined |
| | Nodes-only | no | no | undefined |
| | BART edges-only | yes | no | 0.80 |
| | **BART-role** | **yes** | **yes** | **0.96** |
| | **BART-rel** | **yes** | **yes** | **0.96** |
| | **BART-full** | **yes** | **yes** | **0.96** |

### V. Simulation 4: Model Comparisons for Systematicity and Compatibility (Relational Shift)

We simulated findings from a classic developmental study by Gentner and Toupin (1986). Both source and target analogs involved three animal characters with manipulations of systematicity and compatibility between animals involved in source and target stories. In the simulation, we ran the model for 8 different sets of three animals in the original stories in Gentner and Toupin experiment. (One additional set was omitted because Word2vec representations for a keyword were not available.) These materials are included in an inventory of verbal analogy problems (Ichien, Lu, & Holyoak, 2020). The reported results are from the mappings averaged across the 8 datasets. For each set, we tested all possible assignments of potential mappings for the three animals for the medium compatibility (6) and low compatibility (2) conditions. The high compatibility condition allowed only one possible assignment. Model predictions for all conditions were based on the mean correct mappings for the three animals across all sets and possible assignments.

*Table S5.* Example model comparisons for Simulation 4 (Gentner & Toupin, 1986). Mean number of correct mappings (max = 3) is reported for each condition (older children only).

| Mapping Algorithm | Relation representations | | high compatibility | med compatibility | low compatibility |
|---|---|---|---|---|---|
| PAM | Nodes-only | Systematic | 1.87 | 1 | 0.56 |
| | | Nonsystematic | 1.87 | 1 | 0.56 |
| | BART edges-only | Systematic | 3 | 3 | 3 |
| | | Nonsystematic | 2.25 | 1.44 | 0.94 |
| | **BART-full** | Systematic | **3** | **3** | **2.75** |
| | | Nonsystematic | **3** | **1.31** | **0.94** |

We tested all the usual variants of the PAM model (see Table S2). All the models that include both nodes and edges (i.e., all models except Nodes-only and BART edges-only) can qualitatively account for the basic relational shift (by assuming $\alpha = 2$ for younger children and $\alpha = 1$ for older children). To illustrate predictions for the variables of systematicity and compatibility, Table S5 shows the predictions derived from standard PAM and two other example models using $\alpha = 1$ (i.e., simulating mapping accuracy for the older group). The nodes-only model fails to predict the effect of systematicity (since it lacks relations). At the other extreme, the BART edges-only model predicts perfect accuracy for the systematic condition (since lexical similarity has been eliminated as a factor). The standard PAM model, which is sensitive to both lexical and relational similarity (i.e., both nodes and edges) correctly predicts the interaction between systematicity and compatibility observed in the children's data (see Figure 7 in main paper).

### VI. Simulation 5: Model Comparisons for Solving Convergence Analogy, and Texts

For Simulation 5 (Gick & Holyoak, 1980), we tested all the usual variants of the PAM algorithm, and for each obtained the number of correct mappings for the two story analogs (see texts and extracted keywords in Table S7). Model comparisons are presented in Table S6. Both the standard PAM model and the BART-role variant generated all 7 correct mappings, and the BART-rel variant generated 6 of 7 correct. The other variants were notably less accurate. Thus only the models that include some or all of the full BART relation vectors can solve the convergence analogy reliably.

*Table S6.* Model comparisons for Simulation 5 (Gick & Holyoak, 1980). The number of correct mappings is reported for each model.

| Mapping Algorithm | Relation representations | Correct mappings (max = 7) |
|---|---|---|
| PAM | Word2vec-diff | 2 |
| | Nodes-only | 4 |
| | BART edges-only | 3 |
| | **BART-role** | **7** |
| | BART-rel | 6 |
| | **BART-full** | **7** |

*Table S7.* Full texts of convergence analogy between *The General* story and the radiation problem (Gick & Holyoak, 1980).

### The General (Source)

A small country fell under the iron rule of a dictator. The dictator ruled the country from a strong fortress. The fortress was situated in the middle of the country, surrounded by farms and villages. Many roads radiated outward from the fortress like spokes on a wheel. A great general arose who raised a large army at the border and vowed to capture the fortress and free the country of the dictator. The general knew that if his entire army could attack the fortress at once it could be captured. His troops were poised at the head of one of the roads leading to the fortress, ready to attack. However, a spy brought the general a disturbing report. The ruthless dictator had planted mines on each of the roads. The mines were set so that small bodies of men could pass over them safely, since the dictator needed to be able to move troops and workers to and from the fortress. However, any large force would detonate the mines. Not only would this blow up the road and render it impassable, but the dictator would then destroy many villages in retaliation. A full-scale direct attack on the fortress therefore appeared impossible.

The general, however, was undaunted. He divided his army up into small groups and dispatched each group to the head of a different road. When all was ready he gave the signal, and each group charged down a different road. All of the small groups passed safely over the mines, and the army then attacked the fortress in full strength. In this way, the general was able to capture the fortress and overthrow the dictator.

### Radiation Problem (Target)

Suppose you are a doctor faced with a patient who has a malignant tumor in his stomach. It is impossible to operate on the patient, but unless the tumor is destroyed, the patient will die. There is a kind of ray that can be used to destroy the tumor. If the rays reach the tumor all at once at a sufficiently high intensity, the tumor will be destroyed. Unfortunately, at this intensity, the healthy tissue that the rays pass through on the way to the tumor will also be destroyed. At lower intensities, the rays are harmless to healthy tissue, but they will not affect the tumor either. What type of procedure might be used to destroy the tumor with the rays, and at the same time avoid destroying the healthy tissue?